\documentclass[letterpaper, 10 pt, conference]{ieeeconf}

\IEEEoverridecommandlockouts                              % This command is only needed if 
\usepackage{amsthm}

\usepackage{multicol}
\usepackage{svg}
\usepackage{import}
\usepackage[bookmarks=true]{hyperref}
\usepackage{dsfont}
\usepackage{amsmath}
\usepackage{amsfonts,amssymb}
\usepackage[font=small]{caption}
\usepackage[font=small]{subcaption}
\usepackage{graphicx,wrapfig}
\usepackage{bm}
\newcommand*{\xhat}[1]{#1\kern-0.35em\hat{\phantom{#1}}}
\usepackage[ruled,longend]{algorithm2e}
\usepackage{mathrsfs}
\usepackage{amstext}
\usepackage{float}
\usepackage{nomencl}
\newcommand{\T}{\ensuremath{^\mathsf{T}}}
\usepackage{comment}
\newtheorem{theorem}{Theorem}[section]
\newtheorem{proposition}{Proposition}[section]

\newtheorem{assumption}{Assumption}
\newtheorem{definition}{Definition}
\newtheorem{remark}{Remark}[section]
\newtheorem{lemma}{Lemma}[section]
\usepackage{hyperref}    
 \hypersetup{
     pdfstartview={FitH}, breaklinks=true,
    colorlinks=true, linkcolor=black,
     citecolor=black, filecolor=black,
     urlcolor=black,
 }
\usepackage[capitalise]{cleveref}
\usepackage{xcolor}
\title{An Information-state based Approach to the Optimal Output Feedback Control of Nonlinear Systems}
\author{Raman Goyal$^{*}$, Ran Wang$^{*}$, Mohamed Naveed Gul Mohamed, Aayushman Sharma, Suman Chakravorty % <-this % stops a space
\thanks{$^*$Equal Contribution. The authors are with the Department of Aerospace Engineering, Texas A\&M University, College Station, TX 77843, USA. \{\tt ramaniitrgoyal92, rwang0417, naveed, aayushmansharma, schakrav\}@tamu.edu
}}

\begin{document}

\maketitle

\begin{abstract}
This paper develops a data-based approach to the closed-loop output feedback control of nonlinear dynamical systems with a partial nonlinear observation model. We propose an ``information-state" based approach to rigorously transform the partially observed problem into a fully observed problem where the information-state consists of the past several observations and control inputs. We further show the equivalence of the transformed and the initial partially observed optimal control problems and provide the conditions to solve for the deterministic optimal solution. We develop a data-based generalization of the iterative Linear Quadratic Regulator (iLQR) to partially-observed systems using a local linear time-varying model of the information-state dynamics approximated by an Autoregressive–moving-average (ARMA) model, that is generated using only the input-output data. This open-loop trajectory optimization solution is then used to design a local feedback control law, and the composite law then provides an optimum solution to the partially observed feedback design problem. The efficacy of the developed method is shown by controlling complex high dimensional nonlinear dynamical systems in the presence of model and sensing uncertainty.
\end{abstract} 

% \begin{keywords}
% Partial-state observation, learning and control, optimal control, data-based control, PDE Control, robotics.
% \end{keywords}

% \IEEEpeerreviewmaketitle
% \vspace{-1mm}
\section{Introduction}
% \vspace{-1mm}
The problem of optimal control for nonlinear systems with partial observation finds several applications and is still considered an unsolved problem \cite{roadmap}. It is computationally intractable for complex high-order systems due to the `curse of dimensionality' associated with solving dynamic programming \cite{Bert05}. The problem becomes more challenging when only some of the states are available for measurement, i.e., under partial state observation, which tends to be the case for most of the problems. One approach to solving partially observed problems is to use a feedback policy that acts on the current output, known as output feedback control \cite[Ch.8]{lewis2012optimal}. The dynamic output feedback is a generalization of the static output feedback and also encompasses the approach where a nonlinear observer, such as a high gain observer, extended Kalman filter, or a particle filter, is used to reconstruct the full state of the system, which is then used for the full state feedback control \cite{lewis2012optimal,khalil_outputfb1996}. However, because of the estimation error, output feedback control is often suboptimal compared to full state feedback control, can become very challenging for complex nonlinear systems, and the separation principle may also not hold \cite{lewis2012optimal,khalil_outputfb1996}. Additionally, the problem becomes even more formidable when the model of the system is unknown \cite{Bert05}.
%  This is particularly due to the complex parametrization of the global feedback policy and searching for the policy in this large space has performance issues associated with high variance and reproducibility \cite{henderson2018deep}. %  Many new algorithms that show promising performance are proposed \cite{acktr,trpo,ppo}, and various improvements and innovations have been continuously developed. 
% Reinforcement Learning methods can be divided into model-based and model-free approaches, where a model-based approach uses the system model to find the optimal policy and model-free approach bypasses the need to generate a model to find the optimal policy to control a discrete-time stochastic Markov Decision Process (MDP) \cite{bertsekas2012dynamic,sutton2018reinforcement}. 

In this work, we propose a data-based approach for learning to optimally control complex partially observed nonlinear dynamical systems. The primary idea is to convert the partially observed optimal control problem to a ``fully-observed" problem described using the \textit{information-state} \cite[Ch.5]{Bert05} and to show that the problem described using the information-state is equivalent to the original problem. Then, we use a data-based approach to solve the information state optimal control problem. 
% \noindent \underline{{\textbf{Related Work:}}}
% \mxx{Partially observed problems can be controlled by finding a feedback policy that acts on the current output (Ch.8 \cite{lewis2012optimal}). But these policies, in general, would be suboptimal compared to having a feedback policy acting on the full state. Another approach would be to have a nonlinear observer such as an extended Kalman filter or a particle filter to reconstruct the full state of the system, which can be used for feedback. But, in general, the separation principle doesn't apply for nonlinear systems \cite{robertsson1999observer}. The problem is even more complex for systems whose model is unknown.}

There has been a significant body of work in the field of \textit{learning to control} unknown dynamical systems through Approximate Dynamic Programming (ADP) and Reinforcement Learning (RL) techniques \cite{mnih2015human,silver2016mastering}. These approaches have been successfully used in many areas like playing games \cite{silver2016mastering}, locomotion \cite{lillicrap2015continuous}, and robotic hand manipulation \cite{levine2016end}. However, despite excellent performance on several tasks, reinforcement learning (RL) is still considered very data-intensive, with typically a really large training time. Approaches that consider information state to solve partially observed problems are limited to either linear system models \cite{lewis_info_state,chakrabortty_rl_2020} or finite state and action spaces \cite{ais_jmlr}. There has been little to no work on solving these complex nonlinear problems using output feedback or partial observation models.

% The proposed approach first transforms the partially observed problem into a fully observed problem and then provides the optimal solution to the transformed nonlinear control problem (\mxx{redundant?}). 
In order to find the optimal solution,  the paper introduces a generalization of the iLQR algorithm \cite{ilqg2} that can handle partially observed problems. 
The standard iLQR is a ``local” trajectory-based method, similar to Differential Dynamic Programming (DDP), but only uses first-order dynamics information as opposed to second-order derivatives of the system dynamics needed in DDP \cite{ddp}. 
% Researchers have developed an RL-based iLQR which develops a neural network model from the measurement data and searches for the optimal policy by iteratively refining the neural network \cite{2020neuraliLQR}.
This work suitably generalizes the iLQR method to partially observed systems in a systematic fashion and provides rigorous theoretical justifications for the same. The proposed approach iteratively generates Linear Time-Varying (LTV) state-space models, represented in the information-state, to obtain an optimized nominal information space trajectory. These LTV models are constructed using Autoregressive–Moving-Average (ARMA) models \cite{box2011time} along a nominal trajectory with the input-output perturbation data (rollouts of the system).

The proposed approach is a generalization of the so-called decoupled data-based control (D2C) approach \cite{wang2021decoupled} for designing a feedback controller, to partially observed systems.
%The D2C algorithm introduced a rigorous decoupling of the open-loop (planning) problem from the closed-loop (feedback control) problem. This decoupling allowed for a highly sample-efficient approach to solve the problem in a completely data-based fashion.
% by: first, optimizing the nominal open-loop trajectory of the system using a blackbox simulation model, and then by designing an LQR controller for the identified linearized system. 
%It was shown that the performance of the D2C algorithm is near-optimal to second order in a suitably defined noise parameter \cite{D2C1.0}.
% The D2C approach was used for partially observed systems where the open-loop optimization problem is solved using a general nonlinear programming solver, and a local closed-loop feedback design was generated using an LTV system modeled with Time-Varying Eigensystem Realization Algorithm (TV-ERA) \cite{D2C_CDC19}.
%The Time-Varying Eigensystem Realization Algorithm (TV-ERA) has been used to generate a local feedback law for the D2C approach with a general nonlinear programming solver for the open-loop design \cite{D2C_CDC19}. 
In a recent paper, we proposed an information-state-based system identification using an ARMA model, to generate a local closed-loop feedback controller  \cite{Wang_ICRA_2021}.\\
\underline{\textbf{Contributions:}} This paper builds on the reference \cite{Wang_ICRA_2021} as follows: 1) it provides the theoretical justification of the information-state approach to partially observed problems (Section~\ref{s:InformationState}), 2) it shows that the information-state-based optimal control problem is equivalent to the true partially observed optimal control problem and proves a minimum principle for such problems (Sec.~\ref{s:minimum_principle}), 3) it derives the characteristic equations for solving dynamic programming and shows that there is a unique solution under some assumptions (Sec.~\ref{s:global_optimum}), 4) the minimum principle allows us to generalize the iLQR approach to partially observed problems using the LTV-ARMA identification approach to solve the open-loop nonlinear optimization problem along with the local linear feedback (Sec.~\ref{s:PODiLQR}), which is shown to be highly efficient (20-100x) compared to the gradient descent approach used in \cite{Wang_ICRA_2021}, and 5) it allows us to generate an optimal closed-loop feedback design under noise by augmenting the feedback law with a Kalman filter in the information-state (Sec.~\ref{s:POD2c}). %This LTV system identification approach based on the information-state is critical since existing LTV identification techniques tend to be very brittle and do not scale to the complex nonlinear problems considered in this paper \cite{tv_era}.
% \textcolor{red}{give TVERA ref?}.
% In this paper, we propose an extension of the D2C algorithm which only needs partial observations and not full state information. 
% This allows for the development of a \textit{specific LQG} controller.

The rest of the paper is detailed as follows: 
Section II provides the optimal control problem formulation for the nonlinear system. Section III provides the results to transform a nonlinear partially observed problem into a fully observed problem in an information-state and then provides the equivalence of the optimal control problems along with the conditions to solve for the deterministic global optimal solution. Section IV then gives the framework to solve the problem using information-state-iLQR where the LTV-ARMA model is developed in a data-based fashion.
%The condition on the order of the ARMA model to exactly match the linear system data and the formulation to write a linear time-varying system using \textit{information-state} is also given in section III. 
Section V gives the details of the POD2C closed-loop feedback control algorithm. Finally, empirical results are shown for the partially observed control of complex robotic systems in the presence of process and sensor noise.

%%%%%%%%%%%%%%%%%%%%%%%%%%%%%%%%%%%%%%%%%%%%%%%%%%%%%%%%%%%%%%%%%%%%%%%%%%%%%%%%%%%%%%%%%%%%%%%%%%%%%%%%%%%%
% \vspace{-1mm}
\section{Problem Formulation}
% \vspace{-1mm}
Consider a nonlinear discrete-time dynamical system:
\begin{align}\label{eq:system}
x_{k+1} = f(x_k) + g(x_k)u_k, \quad
z_{k} = h(x_k),
\end{align}
where $x_k \in \mathbb{R}^{n_x}$ is the state, $u_k \in \mathbb{R}^{n_u}$ is the control input and $z_k \in \mathbb{R}^{n_z}$ is the output of the system defined $\forall ~k \geq 0$. The objective of this work is to find the optimal control inputs $\{u_0, u_1, \cdots, u_{N-1}\}$ that minimizes the cost
\begin{align}\label{eq:OCP_PF}
J(x_0) = \sum_{k=0}^{N-1}c(z_k,u_k) + c_N(z_N),
\end{align}
subject to the system model in Eq.~\eqref{eq:system}, where $c(z_k,u_k)$ denotes a running incremental cost and $c_N(z_N)$ denotes a terminal cost function. The goal is to find an output feedback policy, i.e., a policy that only has access to the observations $z_k$, such that the cost above is minimized from an unobserved initial state $x_0$.
% The above formulation shall be termed the partially observed optimal control problem in the rest of the paper.
% The objective of the control design is then to design a feedback policy $u_k(x_k)$ that minimizes the cost function above and is given by:
% \begin{equation}
%     J^*(x_0) = \min_{u_k(\cdot)}E\left[\sum_{k=0}^{N-1}c(x_k,u_k) + \phi(x_k)\right].
% \end{equation}

%%%%%%%%%%%%%%%%%%%%%%%%%%%%%%%%%%%%%%%%%%%%%%%%%%%%%%%%%%%%%%
% \vspace{-1mm}
\section{Information-State Based Control Problem }\label{s:InformationState}
% \vspace{-1mm}
In this section, we introduce the information state and discuss how to transform the original optimal control problem in Eq.~\eqref{eq:OCP_PF} to the information-state domain.

Let $f^n(x_{k-q};u_{k-q},u_{k-q+1},\cdots,u_{k-q+n-1})$ denote the map from the state at time $k-q$, $x_{k-q}$, to the state $x_{k-q+n}$ at time $k-q+n$. Given the initial state and inputs from time $k-q$ to time $k-1$, i.e, $\{x_{k-q};u_{k-q},u_{k-q+1},\cdots,u_{k-1}\}$, we can write the following expressions for the observations $\{z_{k-q},z_{k-q+1},\cdots,z_{k}\}$ as:
% \vspace{-3mm}
\begin{align}
z_{k-q} &= h(x_{k-q}), \nonumber \\
z_{k-q+1} &= h(f^1(x_{k-q};u_{k-q})), \nonumber\\
\nonumber & ~~~   \vdots \\
    z_k &= h(f^q(x_{k-q};u_{k-q},u_{k-q+1},\cdots,u_{k-1})). \label{xt-q}
\end{align}
In the partially observed problem, one can find the underlying state $x_{k-q}$, which is the solution to the above set of nonlinear equations. %can be calculated by posing the following optimization problem:
%\begin{align}
 %   \hat{x}_{k-q} = \arg \min_{x_{k-q}} \|Z_k^q - H_k^q(x_{k-q}) \|, \label{eq:xhatOpt}
%\end{align}
Let $Z_k^q = [z_{k-q}\T,z_{k-q+1}\T,\cdots,z_k\T]\T$, $U_k^q = [u_{k-q}\T,u_{k-q+1}\T,\cdots,u_{k-1}\T]\T$ and $H_k^q(x_{k-q}) = [h(x_{k-q})\T,\cdots, h(f^q(x_{k-q};u_{k-q},\cdots,u_{k-1}))\T]\T$.

\begin{assumption}\label{assump.observability}
\textit{Observability:}
Assume that there exists a finite $\bar{q}$, such that for all $q\geq \bar{q}$, Eq. \ref{xt-q} has a unique solution for $x_{k-q}$, regardless of $(Z_k^q,U_k^q)$.
\end{assumption}
This assumption essentially means that the initial state can be unambiguously reconstructed from a finite history of measurements and control inputs. However, note that this condition needs only be satisfied in principle, and we do not ever explicitly try to solve this equation in our control synthesis.
Due to the implicit function theorem and the observability assumption, we can write $x_{k-q}$ as some unique function $\bar{f}(.,.)$ of past measurements and inputs as:
\begin{align}
    {x}_{k-q} = \bar{f}(Z_k^q,U_k^q). \label{eq:xhat_past}  
\end{align}
\begin{remark}\label{remark.III.1}
    Note that typically for a partially observed problem, one would use a (nonlinear) observer to estimate the state, which would then be used to specify the control action assuming the estimated state to be the true state (certainty equivalence). Owing to the observability assumption \ref{assump.observability}, given the first $\bar{q}$ inputs and outputs, one can, in principle, exactly reconstruct the initial state, and thus predict the state evolution exactly after $\bar{q}$ steps, thereby mapping back to the fully observed problem. So, we shall assume that the first $\bar{q}$ inputs are specified in the partially observed control formulation we discuss later in this section. Finally, note that a typical nonlinear observer can never perfectly reconstruct the state in a finite number of steps.
\end{remark}
%Notice that the true initial condition ${x}_{k-q}$ is the global minimum of the above problem. 

Next, let us write the state at current time $x_k$ as:
$x_k = f^q(x_{k-q};u_{k-q},u_{k-q+1},\cdots,u_{k-1})$, which can be written again by substituting for ${x}_{k-q}$ from \cref{eq:xhat_past} in some unique functional form, based on observability assumption and implicit function theorem, as:
$%$\begin{align}
 x_k = \Psi(Z_k^q,U_k^q). ~
$%\end{align}
Let us now finally define the ``\textit{Information-State}".
\begin{definition} 
\textbf{Information-state}. The information-state \cite[Ch.5]{Bert05}, \cite[Ch.6]{kumar1986stochastic} of the system in Eq.~\eqref{eq:system} of order $``q"$ (at time $k$) is defined as 
$
\mathcal{Z}_k^q = \begin{bmatrix}
    z_k\T,z_{k-1}\T,\cdots,z_{k-q}\T,u_{k-1}\T,\cdots,u_{k-q}\T    \end{bmatrix}\T  \in \mathbb{R}^n,
$
where $n = (q+1)n_z + qn_u$.
\end{definition}
The above definition allows us to write: $x_k = \Psi(\mathcal{Z}_k^q)$. Further, due to the implicit function theorem, if the dynamics and the observation functions $f$, $g$, and $h$ are $\mathcal{C}^k$, so is the map $\Psi$.
The above development can be summarized as:
%\vspace{-1mm}
\begin{lemma}\label{L:Information}
Given Assumption 1, there exists a unique function $\Psi(\cdot)$, such that the state at time $k$, $x_k = \Psi (\mathcal{Z}_k^q)$. In particular, if $f(\cdot)$ and $h(\cdot)$ are $\mathcal{C}^k$, so is the function $\Psi$.
\end{lemma}
\noindent The above result shows that the state at time $k$ is some nonlinear map of the observation and the control inputs at the previous $``q"$ time steps. 
% \mxx{
% \begin{lemma} \label{lemma.unique transformation}
%    Given $x_k$ and the input-output trajectory from the finite past: $\{z_k, z_{k-1}, \cdots, z_{k-q+1}, u_{k-1}, \cdots, u_{k-q+1} \}$, there is a unique $\mathcal{Z}_k^q$.
% \end{lemma}
% This result arises from the definition of the information-state. \todo{Is this needed?}
% }

The remainder of this section shows how to transform the optimal control problem into the information-state domain. Consider the observation of the system as: 
% \begin{align*}
$    z_{k} = h(x_k) = h(f(x_{k-1}) + g(x_{k-1})u_{k-1}). $
% \end{align*}
Using the relationship, $x_{k-1} = \Psi (\mathcal{Z}_{k-1}^q)$, the observation can be written in the form, $z_k = \tilde{h}(\mathcal{Z}_{k-1}^q, u_{k-1})$, and further the equation can be written in terms of Information-state as:
$
    \mathcal{Z}_{k}^q = \mathcal{H}(\mathcal{Z}_{k-1}^q, u_{k-1}).
$

This can be summarized in the following result.
\begin{lemma}\label{lemma.input_output_response}
Under Assumption~\ref{assump.observability}, the systems $x_{k+1} = f(x_k) + g(x_k) u_k , z_k  = h(x_k)$ and $ \mathcal{Z}^q_{k+1} = \mathcal{H}(\mathcal{Z}^q_k, u_k)$  have the same input-output response after $k \geq q$, regardless of the ``unknown" initial condition $x_0$.
\end{lemma}
% \mxx{\begin{proof}
% We can define the output of the information-state from Eq.~\eqref{eq.info-state model} as, 
% \begin{align*}
% z'_k & = \underbrace{[I_{m \times m}, 0, \cdots, 0]}_{\mathcal{C}} \mathcal{Z}^q_k.
% \end{align*}
% From the definition of $\mathcal{Z}^q_k$, $z'_k = z_k$.
% \end{proof}
% }
Recall $\bar{q}$ is the minimum number of observations required to satisfy Assumption~\ref{assump.observability}. Hence, the information state model needs at least $\bar{q}$ measurements and inputs before it can start prediction, as these are required to form the initial information state. We assume that the first $\bar{q}$ control inputs and resulting observations are specified to us, which implies that the state $x_{\bar{q}}$ is known given the observability assumption. Then, we may reformulate the original partially-observed control problem as the following fully observed problem starting at the initial state $x_{\bar{q}}$:
%\vspace{-1mm}
\begin{subequations}\label{eq:OptProbX}
   \begin{align}
  J_x^*  &= \min_{u_k}  \sum^{N-1}_{k=\bar{q}} c(z_k, u_k) + c_N(z_N), \\
  \text{s.t.} ~~ x_{k+1} &= f(x_k) + g(x_k) u_k,~ \text{given} ~ x_{\bar{q}}. \label{eq:OptProbX_dynamics}
\end{align} 
\end{subequations}

Further, we pose the following ``fully observed" optimal control problem in terms of the Information-State assuming that the same initial $\bar{q}$ inputs are applied as above:
% \vspace{-1mm}
\begin{subequations}\label{eq:OptProbZ}
\begin{align}
    J_{\mathcal{Z}}^* = \min_{u_k}  \sum^{N-1}_{k=\bar{q}} c(z_k, u_k) + c_N(z_N), \\
    \text{s.t.} ~~ 
    \mathcal{Z}^q_{k+1} = \mathcal{H}(\mathcal{Z}^q_k, u_k), ~\text{given}~\mathcal{Z}^{\bar{q}}_{\bar{q}}.\label{eq:OptProbZ_dynamics}
\end{align}
\end{subequations}

Hence, we have transformed our optimal control problem to the information-state domain in Eq.~\eqref{eq:OptProbZ}. 
%The next section discusses the optimality of this information-state problem and also discusses the control inputs that should be chosen for the first $\bar{q}$ time steps.

%\vspace{-1mm}
\subsection{Optimality of the Information-State Control Problem}\label{s:optimality_info_state}

This section discusses the equivalence of solving the optimal control problem using the information-state dynamics and the partially observed state-space dynamics and proves that transforming the problem to the information-state leads to no loss in optimality after a finite initial transience.
The equivalence of the solution to the two problems is established in the following result.

\begin{theorem}\label{theorem.optimal_fb}
    Given the initial information-state $\mathcal{Z}^{\bar{q}}_{\bar{q}}$ or equivalently the state $x_{\bar{q}}$, the solution to the optimal control problem (Eq.~\eqref{eq:OptProbX}) is identical to the solution to the optimal information state control problem (Eq.~\eqref{eq:OptProbZ}).
\end{theorem}
\begin{proof}
Consider the two problems in Eqs.~\eqref{eq:OptProbX} and \eqref{eq:OptProbZ}.
To show the equivalence between the two problems, one needs to show: (i) the initial conditions are equivalent; (ii) the input-output response of the state-space system and information-state system are equal; (iii) the optimal feedback policies of both the systems generate the same control inputs.

Using \cref{L:Information}, the initial condition for the systems are related by $x_{\bar{q}} = \Psi(\mathcal{Z}^{\bar{q}}_{\bar{q}})$, where the information-state $\mathcal{Z}^{\bar{q}}_{\bar{q}}= [z_{\bar{q}}\T,\cdots,z_0\T,u_{\bar{q}-1}\T,\cdots,u_0\T]\T$. Further, Lemma~\ref{lemma.input_output_response} shows they have the same input-output response.

To show that the feedback policies generate the same control inputs,  let the optimal control policy mapping the state to the control at time $k$ for the state-space and information-space system be $\pi^k_x(\cdot)$, and $\pi_{\mathcal{Z}}^k(\cdot)$, respectively.     
The policy for the true full-state feedback $\pi_x^k(\cdot)$ can be applied to the information-state system, by using the transformation from \cref{L:Information}, $x_k = \Psi(\mathcal{Z}_{k})$, which gives the control input to be  $u_k = \pi_x^k(\Psi( \mathcal{Z}_{k}))$. Then, 
because of optimality of the policy $\pi_{\mathcal{Z}}(\cdot)$ for the information-state system, the input-output equivalence of the systems in Eqs.~\eqref{eq:OptProbX_dynamics} and~\eqref{eq:OptProbZ_dynamics}, and the same cost functions, we get:
 $
    J_{\mathcal{Z}}^{\pi_{\mathcal{Z}}} \leq J_x^{\pi_x}.
$
Similarly, the optimal information-state policy  $\pi_{\mathcal{Z}}(\cdot)$, can be applied to the true state-space model by taking the past inputs and outputs as the argument to the policy. Because of optimality of $\pi_x$ for the state space system, we get: 
$
    J_x^{\pi_x} \leq J_{\mathcal{Z}}^{\pi_{\mathcal{Z}}}.
$
Owing to the above two inequalities, $J_x^{\pi_x} =J_{\mathcal{Z}}^{\pi_{\mathcal{Z}}} $. Hence, the optimal feedback on the information-state is identical to the optimal control for the underlying state-space system.
\end{proof}

\begin{remark}
Notice that the two problems (Eq.~\eqref{eq:OptProbX} and Eq.~\eqref{eq:OptProbZ}) mention the initial state to be defined at $k = \bar{q}$ and assume that the initial few control inputs $\{u_0, u_1, \cdots, u_{\bar{q}-1}\}$ are specified. Intuitively, one can see that these inputs are ambiguous as they are needed to get enough information to reconstruct the initial state/ form the initial information state and should not be included in the optimization. Thus, it is advisable to take small perturbations to keep the system near the initial state till the state can be reconstructed at $k = \bar{q}$.  Nonetheless, given that $\bar{q} \ll T$, we can expect that this initial transient will not affect the total cost significantly.
\end{remark}
%\todo{
%\begin{remark}
%Please notice that the above two defined problems (Eq.~\eqref{eq.OptProbZ} and Eq.~\eqref{eq.OptProbZ}) mention the initial state to be defined at $t = \bar{q}-1$ and does not mention the initial few control inputs $\{u_0, u_1, \cdots, u_{\bar{q}-2}\}$. Since the state is unknown, it is not clear what the optimal control has to be. It is advised to take small perturbations to keep the system near the initial state till the state can be reconstructed at $t = \bar{q}-1$.  %\end{remark}}

%\vspace{-1mm}
\section{A Minimum Principle for the Information-state based control problem}\label{s:minimum_principle}
%\vspace{-1mm}
In this section, we develop a minimum principle for the information-state optimal control problem. We make the following assumption for the problem given in Eq.~\eqref{eq:OptProbZ} and show that the information-state dynamics is control-affine. 
\begin{assumption} \label{assump: quadratic_cost}
    The incremental cost in Eq.~\eqref{eq:OptProbZ} is quadratic in the control: $c(z_k,u_k) = l(z_k) + \frac{1}{2}u_k\T R u_k$, where $R$ is positive definite.
\end{assumption}
\begin{proposition}\label{assump: affine_model}
Given the underlying system dynamics is affine in control with the form: $\dot{x} = f(x) + g(x)u$, the information-state model is affine in control:
%\vspace{-1mm}
\begin{align}\label{eq: info_state_model_affine}
     \mathcal{Z}^q_{k} = \mathcal{F}(\mathcal{Z}_{k-1}^q)+ \mathcal{G}(\mathcal{Z}_{k-1}^q) u_{k-1} .
\end{align}
\end{proposition} 
\begin{proof}
We write the system dynamics with a forward Euler approximation for a small discretization time: 
$x_{k+1} = x_k + f(x_k)  \Delta t + g(x_k) u_k \Delta t$, and the observation model as: $z_{k} = h(x_{k}) =  h(x_{k-1}+d x_k)$, where $dx_k = f(x_{k-1})  \Delta t + g(x_{k-1}) u_{k-1} \Delta t$. Given that the time discretization is sufficiently small, the observation model can be written using a first order Taylor expansion as:
% \begin{align*}
%     z_{k} &= h(x_{k-1}+d x_k) = h(x_{k-1})+ \underbrace{\frac{\partial h}{\partial x}\Big|_{x_{k-1}}}_{H(x_{k-1})} d x_k,\\
%     % z_{k} &= h(x_{k-1})+ H(x_{k-1}) d x_k,\\
%     z_{k} &= \underbrace{h(x_{k-1})+ H(x_{k-1}) f(x_{k-1})  \Delta t}_{F(x_{k-1})} +\\
%     & ~~~~~\underbrace{H(x_{k-1}) g(x_{k-1}) \Delta t}_{G(x_{k-1})} u_{k-1}.
% \end{align*}
\begin{align*}
 z_{k} &= h(x_{k-1}+d x_k) = h(x_{k-1})+ \underbrace{\frac{\partial h}{\partial x}\Big|_{x_{k-1}}}_{H(x_{k-1})} d x_k,\\
z_{k}& = \underbrace{h(x_{k-1})+ H(x_{k-1}) f(x_{k-1})  \Delta t}_{F(x_{k-1})} + \\
 &\quad \underbrace{H(x_{k-1}) g(x_{k-1}) \Delta t}_{G(x_{k-1})} u_{k-1}.
\end{align*}

Thus, we can write the observation $z_k$ as some function of $x_{k-1}$ and $u_{k-1}$ as:
% \begin{align}
$     z_{k} = F(x_{k-1})+ G(x_{k-1}) u_{k-1} .$
% \end{align}
Further substituting for $x_{k-1}$ as $x_{k-1} = \Psi(\mathcal{Z}_{k-1}^q)$ from above, we can write the $z_k$ in terms of information-state as:
\begin{align}\label{eq.info_output}
     z_{k} = F(\Psi(\mathcal{Z}_{k-1}^q))+ G(\Psi(\mathcal{Z}_{k-1}^q)) u_{k-1},
\end{align}
and finally, the entire equation can trivially be written in terms of $\mathcal{Z}^q_{k}$ in a control affine form as in \cref{eq: info_state_model_affine}. 
\end{proof}

Using the above results and considering the time-index $k=0$ to be the first time index the information-state in constructed, the problem in Eq.~\eqref{eq:OptProbZ} can be rewritten as
\begin{subequations}\label{eq:OptProbZ_1}
    \begin{align}
    J = \min_{u_k}  \sum^{N-1}_{k=0} (l(z_k) + \frac{1}{2}u_k\T R u_k) + c_N(z_N), \\
    \text{s.t.} ~~ 
    \mathcal{Z}_{k+1} =\mathcal{F}(\mathcal{Z}_{k})+ \mathcal{G}(\mathcal{Z}_{k}) u_{k} , ~\text{given}~\mathcal{Z}_{0}.   
\end{align}
\end{subequations}
 
The superscript $q$ in $\mathcal{Z}^q$ and subscript $\mathcal{Z}$ in $J_{\mathcal{Z}}$ are ignored for the sake of convenience.
The following result gives the necessary conditions for the solution $\{\bar{\mathcal{Z}_k}, \bar{u}_k \}$ to the above fully observed problem in \cref{eq:OptProbZ_1} has to satisfy.
\begin{theorem}\label{theorem:InformState_optimality}
Let the cost functions $l(\cdot)$, $c_N(\cdot)$, the drift $\mathcal{F}(\cdot)$ and the input influence function $\mathcal{G}(\cdot)$ be $\mathcal{C}^2$, i.e., twice continuously differentiable. 
The minimum of the open-loop problem (Eq.~\eqref{eq:OptProbZ_1}) starting at some initial information-state $\mathcal{Z}_0$ satisfies:
\begin{align}
    &\bar{u}_k = - R^{-1} \mathcal{G}(\bar{\mathcal{Z}}_k)\T G_{k+1}, \label{eq:NC_control} \\
    &\mathcal{\bar{Z}}_{k+1} = \mathcal{F}(\bar{\mathcal{Z}}_k) - \mathcal{G}(\bar{\mathcal{Z}}_k) R^{-1} \mathcal{G}(\bar{\mathcal{Z}}_k)\T G_{k+1} \label{eq:NC_state}, \\
     &G_k = \bar{L}_k^\mathcal{Z} + \mathcal{A}_k\T G_{k+1},  \label{eq:MP} \\
     \nonumber &P_k = \mathcal{A}_k\T P_{k+1} \mathcal{A}_k + \bar{L}_k^{\mathcal{Z}\mathcal{Z}} + \sum_{i=1}^{n}  [\bar{\mathcal{F}}_{k,i}^{\mathcal{Z}\mathcal{Z}} + \sum_{j=1}^{n_u} \bar{\Gamma}_{k,i}^{j,\mathcal{Z}\mathcal{Z}} \bar{u}_k^j]G_{k+1}^i \label{eq: P_k}\\ 
    &~~~~~~~~~~~~~~~~ - K_k^{T} (R_k + \mathcal{B}_k\T P_{k+1} \mathcal{B}_k) K_k,\\
    \nonumber &K_k = - (R_k + \mathcal{B}_k\T P_{k+1} \mathcal{B}_k)^{-1} [\sum_{i=1}^{n} \bar{\mathcal{G}}_{k,i}^{\mathcal{Z},T}  G_{k+1}^i + \mathcal{B}_k\T P_{k+1} \mathcal{A}_k], %\label{eq:MPK}
% \vspace{-2mm}
\end{align}
where $\mathcal{A}_k = \bar{\mathcal{F}}_k^\mathcal{Z} + \sum_{j=1}^{n_u} \bar{\Gamma}_k^{j,\mathcal{Z}} \bar{u}_k^j, ~\mathcal{B}_k = \mathcal{G}(\bar{\mathcal{Z}}_k)$, and $\{\bar{\mathcal{Z}}_k\}$ represents the optimal nominal trajectory, $\bar{L}_k^\mathcal{Z} = \nabla_\mathcal{Z} l|_{\bar{\mathcal{Z}}_k}$, $G_k = \nabla_\mathcal{Z} J_k|_{\bar{\mathcal{Z}}_k}$, with terminal condition  $G_N = \nabla_{\mathcal{Z}} c_N|_{\bar{\mathcal{Z}}_N}$, $P_k = \nabla_{\mathcal{ZZ}}J_k|_{\bar{\mathcal{Z}}_k}$, with terminal condition $P_N = \nabla_{\mathcal{ZZ}}c_N|_{\bar{\mathcal{Z}}_N}$, $\bar{u}_k =  [\bar{u}_k^1 \cdots \bar{u}_k^{n_u}]\T$, the control influence matrix: $\mathcal{G} =  \begin{bmatrix} \Gamma^1 (\mathcal{Z}) \cdots \Gamma^{n_u} (\mathcal{Z}) \end{bmatrix}$, and $\Gamma^j$ and $\bar{u}_k^j$ represents the control influence vector and optimal control vector corresponding to the $j^{th}$ input. Finally, $\bar{\mathcal{F}}^\mathcal{Z}_k = \nabla_\mathcal{Z} \mathcal{F}|_{\bar{\mathcal{Z}}_k}$ and  $\bar{\Gamma}^{j,\mathcal{Z}}_k = \nabla_\mathcal{Z} \Gamma^j|_{\bar{\mathcal{Z}}_k}$ gives the Jacobians of the system dynamics and $\bar{\mathcal{F}}_{k,i}^{\mathcal{Z}\mathcal{Z}}$ and  $\bar{\Gamma}_{k,i}^{j,\mathcal{Z}\mathcal{Z}}$ gives the Hessians of the system dynamics along the nominal trajectory.
\end{theorem}

\begin{proof}
Equations~\eqref{eq:NC_control} to \eqref{eq:MP} are the standard necessary conditions in optimal control for nonlinear systems \cite{lewis2012optimal}. 
The derivation for equations $P_k$ and $K_k$ is given in \cite{wang2022searcharxiv}.
\end{proof}
% \mxx{Should we say the equation for P as also a necessary condition? the optimal control doesn't need that. Also, should we state that P has to be positive definite around the optimal trajectory as a condition, or is it trivial?}

\subsection{A Global Optimum for the Nonlinear Problem}\label{s:global_optimum}
We proceed to show that the problem in \cref{eq:OptProbZ_1} has a unique minimum under some assumptions. 
The problem in \cref{eq:OptProbZ_1} can equivalently be written as the following Dynamic Programming problem:
\begin{align}
    &J_k(\mathcal{Z}_k) = \min_{u_k} \Big[l(z_k) + \frac{1}{2}u_k\T R u_k + J_{k+1}(\mathcal{Z}_{k+1})  \Big], \label{eq:DP}\\
    &\text{where}\ J_N(\mathcal{Z}_N) = c_N(z_N). \nonumber
\end{align}
The minimizing control $u_k$ for the above problem satisfies the first-order necessary condition $\frac{\partial J_k}{\partial u_k} = 0$, and given the current state is $\bar{\mathcal{Z}}_k$, the control is given by
\begin{align}
   \nonumber &\bar{u}_k = - R^{-1} \mathcal{G}(\bar{\mathcal{Z}}_k)\T \frac{\partial J_{k+1}}{\partial \mathcal{Z}_{k+1}}\Big|_{\bar{\mathcal{Z}}_{k+1}} = - R^{-1} \mathcal{G}(\bar{\mathcal{Z}}_k)\T G_{k+1} , 
   % \label{eq:characteristic_eq_control}
   \\
   &\text{where,}~ \bar{\mathcal{Z}}_{k+1} = \mathcal{F}(\bar{\mathcal{Z}}_k) - \mathcal{G}(\bar{\mathcal{Z}}_k) R^{-1}\mathcal{G}(\bar{\mathcal{Z}}_k)\T G_{k+1}. \label{eq:characteristic_eq_state}
\end{align}
Substituting this control in Eq.~\eqref{eq:DP} gives:
\begin{align}
    J_k(\mathcal{Z}_k) = l(z_k) + \frac{1}{2} G_{k+1}\T  \mathcal{G}(\mathcal{Z}_k) R^{-1} \mathcal{G}(\mathcal{Z}_k)\T G_{k+1} + \nonumber\\
    J_{k+1}(\mathcal{F}(\mathcal{Z}_k) - \mathcal{G}(\mathcal{Z}_k) R^{-1}\mathcal{G}(\mathcal{Z}_k)\T G_{k+1})  \label{eq: optimal_cost}.
\end{align}
Taking the partial derivative of Eq.~\eqref{eq: optimal_cost} with respect to $\mathcal{Z}_k$ at $\bar{\mathcal{Z}}_k$  gives rise to the co-state equation: 
\begin{align}
    G_k &= \bar{L}_k^\mathcal{Z} + (\bar{\mathcal{F}}^{\mathcal{Z}}_k - \bar{\mathcal{G}}^{\mathcal{Z}}_k R^{-1}  \mathcal{G}(\bar{\mathcal{Z}}_k)\T G_{k+1})\T G_{k+1} \label{eq:characteristic_eq_costate}.
\end{align}
(Note: In \cref{eq:characteristic_eq_costate}, $\bar{\mathcal{G}}^{\mathcal{Z}}_k$ would be a tensor for the vector case, but we abuse the notation for the sake of illustration.)
Since \cref{eq:DP} is a terminal value problem, we start from the terminal state. For a given terminal condition $\{\bar{\mathcal{Z}}_N$, $\bar{G}_N = \frac{\partial c_N}{\partial \mathcal{Z}_N}|_{\bar{\mathcal{Z}}_N}\}$, \cref{eq:characteristic_eq_state,eq:characteristic_eq_costate} can be solved backward in time to find the optimal solution. Hence, \cref{eq:characteristic_eq_state,eq:characteristic_eq_costate} are the characteristic equations to the solution to the dynamic programming problem in \cref{eq:DP}. But, they need not have a unique backward evolution. 
\begin{remark}
    Equations~\eqref{eq:characteristic_eq_state},~\eqref{eq:characteristic_eq_costate} are the discrete-time analog to the characteristic ODEs one obtains for the Hamilton-Jacobi-Bellman equation using the method of characteristics \cite{mohamed2020optimal}.
\end{remark}
%\textit{Note: It might seem that \cref{eq:characteristic_eq_state,eq:characteristic_eq_costate} are nothing but the necessary conditions for optimal control. Yes, indeed they are. However, they give a new viewpoint from the Dynamic Programming approach to solve terminal value problems and connect them to classical optimal control, which solves them as two-point boundary value problems.}
% For the development in the remainder of this section, the following assumption is made. 
\begin{assumption}\label{assump:unique_x}
Given $\bar{\mathcal{Z}}_{k+1}, \bar{G}_{k+1}$, \cref{eq:characteristic_eq_state} has a unique solution for $\bar{\mathcal{Z}}_{k}:$
$
    \bar{\mathcal{Z}}_{k} = \mathscr{F}(\bar{\mathcal{Z}}_{k+1}, \bar{G}_{k+1}), 
$
where $\mathscr{F}(\cdot,\cdot)$ is an implicit function. Moreover, two different state-costate pairs - $\{\bar{\mathcal{Z}}^{(1)}_{k+1}, \bar{G}^{(1)}_{k+1}\}$ and $\{\bar{\mathcal{Z}}^{(2)}_{k+1}, \bar{G}^{(2)}_{k+1}\}$ - cannot lead to the same $\{\bar{\mathcal{Z}}_{k}, \bar{G}_{k}\}$ under \cref{eq:characteristic_eq_state,eq:characteristic_eq_costate}.
\end{assumption}
A couple of scenarios where the above assumption is satisfied are shown below. 
\begin{lemma}\label{lemma.small_deltat}
    Let the functions, $\mathcal{F}, \mathcal{G}$ be continuously differentiable. If the underlying discretization time for the systems in \cref{eq:characteristic_eq_state,eq:characteristic_eq_costate} are sufficiently small, i.e., $\Delta t \rightarrow 0$, then, there is a unique map from $\{\bar{\mathcal{Z}}_{k+1}, \bar{G}_{k+1} \}$ to $\{\bar{\mathcal{Z}}_{k}, \bar{G}_{k} \}$ and vice-versa.
\end{lemma}
\begin{proof} 
    Since $\Delta t \rightarrow 0 $, the implicit function theorem shows that, in the neighborhood of $\{\bar{\mathcal{Z}}_{k+1}, \bar{G}_{k+1} \}$, there exists a unique $\{\bar{\mathcal{Z}}_{k}, \bar{G}_{k} \}$, that satisfies \cref{eq:characteristic_eq_state,eq:characteristic_eq_costate}. The same argument applies to the forward map.   
\end{proof}
\begin{lemma}\label{lemma.state_independent}
    If the control influence matrix $\mathcal{G}(\cdot)$ is state independent, then, there is a unique map from $\{\bar{\mathcal{Z}}_{k+1}, \bar{G}_{k+1} \}$ to $\{\bar{\mathcal{Z}}_{k}, \bar{G}_{k} \}$ and vice-versa.
\end{lemma}
\begin{proof} 
    Given $\{\bar{\mathcal{Z}}_{k+1}, \bar{G}_{k+1} \}$, $\bar{u}_k$ is exactly determined from \cref{eq:NC_control} without the knowledge of $\bar{\mathcal{Z}}_{k}$, since $\mathcal{G}(\cdot)$ is state independent. Since the state reached by a dynamical system $\bar{\mathcal{Z}}_{k+1}$, and the control action $\bar{u}_k$ taken to reach there are known, the state it originated from $\bar{\mathcal{Z}}_{k}$ is unique. Also, \cref{eq:characteristic_eq_costate} is a linear equation for state independent $\mathcal{G}$; hence $\bar{G}_k$ is unique. The uniqueness of the forward map can be similarly shown. 
\end{proof}
\noindent Now, we show \cref{eq:characteristic_eq_state,eq:characteristic_eq_costate} have a unique backward evolution and also show that trajectories originating from different terminal conditions do not cross. 
% \begin{figure}[!htbp]
%   \centering
%   \includesvg[width=0.65\textwidth]{figures/vec_image.svg}
%   \caption{An illustration of how the state and costate evolve backward in time. The costate remains constant, while the state changes continuously between time-steps.}
%   \label{fig.state_costate_evolution}
% \end{figure}
\begin{figure}
    \centering
    \def\svgwidth{0.9\columnwidth}
    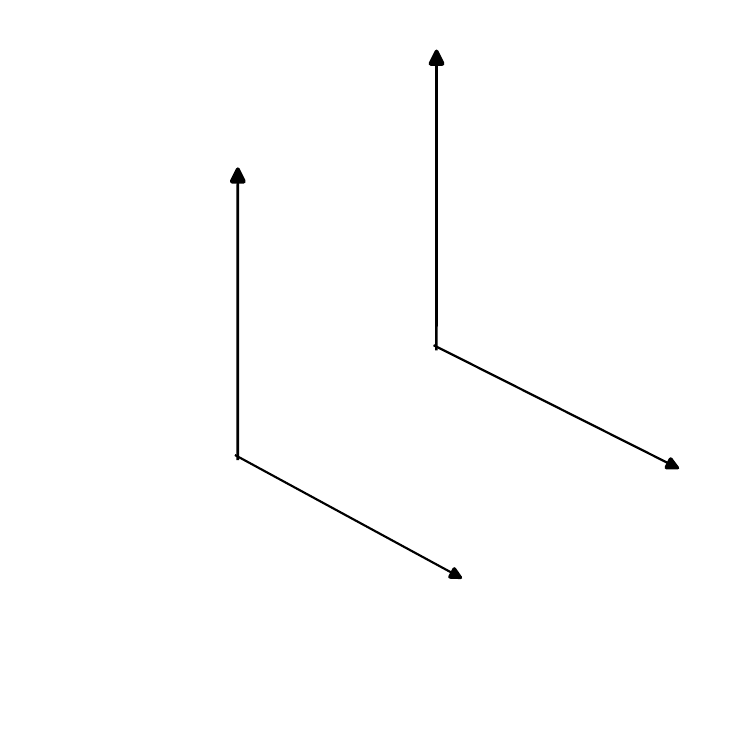
    \caption{An illustration of how the state and costate evolve backward in time. The costate remains constant, while the state changes continuously between time-steps.}
    \label{fig.state_costate_evolution}
\end{figure}
\begin{lemma}\label{lemma.unique_evolution}
    Under Assumption \ref{assump:unique_x}, given terminal conditions $\{\bar{\mathcal{Z}}_N$, $\bar{G}_N =  c^{\mathcal{Z}}_N|_{\bar{\mathcal{Z}}_N}\}$, the state and costate will have a unique evolution backward in time.
\end{lemma}
\begin{proof}
    Let us consider at $k=N-1$. Since $\{\bar{\mathcal{Z}}_N$, $\bar{G}_N\}$ are given, $\bar{\mathcal{Z}}_{N-1}$ is unique due to Assumption~\ref{assump:unique_x}. Then, it is straightforward to see that $\bar{G}_{N-1}$ is unique from \cref{eq:characteristic_eq_costate}. The same argument is applicable at every time step $k$.
\end{proof}    
\begin{lemma}\label{lemma.intersect}
    Under Assumption \ref{assump:unique_x}, trajectories originating from different terminal conditions do not cross in the $\{\mathcal{Z}, G \}$ space for all time.
\end{lemma}
\begin{proof}
    The only scenario the trajectories might cross is in between the time steps. The state of the dynamical system, in practice, will be continuously evolving in between time steps, while the costate will be constant (see Fig.~\ref{fig.state_costate_evolution}). If trajectories have to intersect, they should start with the same costate, but different states. At the crossing point, they will have the same state, and a dynamical system cannot have different evolutions from the same state and costate.
\end{proof}
\begin{figure}[!htbp]
\centering
  \def\svgwidth{\columnwidth}
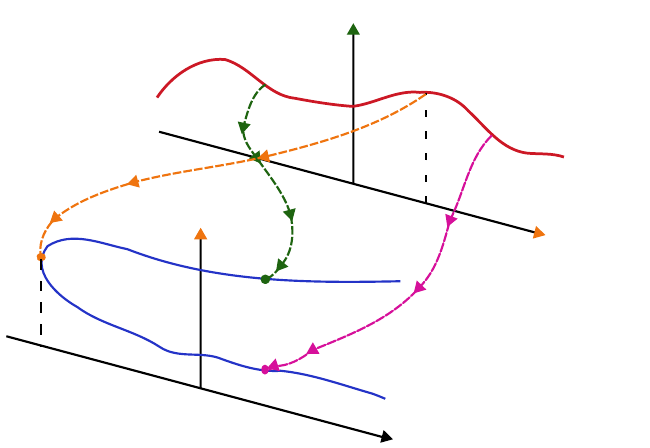
  \caption{Mapping of the terminal conditions under the characteristic equations. For two characteristic curves to flow through the same state $\bar{\mathcal{Z}}_k$, the $\phi_k(.)$ curve has to fold on itself necessitating the existence of a state $\mathcal{Z}_k^*$ such that $\frac{d\mathcal{Z}_k}{dG_k}|_{\mathcal{Z}_k^*} = 0$. Given that the characteristic equations have unique solutions, such a state $\mathcal{Z}_k^*$ cannot exist. }
  \label{fig.state_costate_uniqueness}
\end{figure}
\begin{theorem}\label{OL_optimality}
\textit{Global Optimality of open-loop solution.} Let the cost functions $l(.)$, $c_N(.)$, the drift $\mathcal{F}(.)$ and the input influence function $\mathcal{G}(.)$ be $\mathcal{C}^2$, i.e., twice continuously differentiable. Then, an optimal trajectory that satisfies the Minimum Principle (\cref{theorem:InformState_optimality}) from a given initial state $\mathcal{Z}_0$, is the unique global minimum of the open-loop problem starting at the initial state $\mathcal{Z}_0$.
\end{theorem}
\begin{proof} 
 Let us denote $G_N = c_N^{\mathcal{Z}}(\mathcal{Z}_N) \equiv \phi_N(\mathcal{Z}_N)$. Next, we use \cref{lemma.unique_evolution,lemma.intersect} to show that, under the state and costate equations -\cref{eq:characteristic_eq_state,eq:characteristic_eq_costate}- the function $\phi_N(\mathcal{Z}_N)$ remains a function, i.e., we can write $G_k = \phi_k(\mathcal{Z}_k)$, for some suitable smooth function $\phi_k(.)$, for any $k \in \{0,\cdots,N\}$. Hence, there cannot be multiple costates for the same state (see Fig.~\ref{fig.state_costate_uniqueness}).
Next, note that if a solution satisfies the initial state $\mathcal{Z}_0$, then it means that we have found a terminal state $\mathcal{Z}_N$, along with the terminal co-state $G_N = c_N^\mathcal{Z}(\mathcal{Z}_N)$, that satisfies the necessary conditions. However, this is, by definition, a solution that is found by satisfying the Minimum Principle. Therefore, owing to the development above, the co-state $G_0 = \phi_0(\mathcal{Z}_0)$ is uniquely determined by the initial state $\mathcal{Z}_0$, and a solution that satisfies the minimum principle is necessarily unique. Moreover, since this solution is the unique solution to the DP problem (\cref{eq:DP}) with initial state $\mathcal{Z}_0$, it is also the global optimum.
\end{proof}

\section{Open-Loop Trajectory Design using Partially Observed Data based iLQR (POD-iLQR)}\label{s:PODiLQR}
%\vspace{-1mm}
% The iLQR algorithm solves the nonlinear optimal control problem by iteratively solving the LQR problem.
The ILQR algorithm, in theory, owing to the minimum principle, can be generalized to iteratively solve the nonlinear information-state optimal control problem from section II. 
\begin{remark}
In our recent arxiv paper \cite{wang2022searcharxiv}, we showed the iLQR algorithm converges to the global optimal solution under some mild assumptions on cost functions (quadratic in control) and system dynamics (affine in control). Thus, the iLQR generalization to the proposed information-state based formulation would also result in convergence to the global optimal solution under the same assumptions.
\end{remark}
\noindent \textit{Information-state based generalized iLQR algorithm:}

\noindent \textit{\textbf{Forward Pass -}}
Given a nominal control sequence $\{u_k\}_{k=0}^{N-1}$ and initial information-state vector $\mathcal{Z}_0$, the state is propagated in time in accordance with the dynamics
$%$\begin{equation*}
    \mathcal{Z}_{k+1}=\mathcal{H}(\mathcal{Z}_k, u_k).
$%\end{equation*} 
~Thus, we get the nominal information-state trajectory $(\bar{\mathcal{Z}}_k,\bar{u}_k)$.\\
\textit{\textbf{Local LTV System -}}
Next, we find the corresponding local LTV system around the trajectory $(\bar{\mathcal{Z}}_k,\bar{u}_k)$, which can be written as:
% Given a trajectory $\{x_k\}_{k=0}\T$ corresponding to control input $\{u_k\}_{k=0}^{N-1}$, we can use input-output perturbation and a Linear Least Squares method to estimate the LTV system given as
$%$\begin{equation}
    \delta \mathcal{Z}_{k+1} = \mathcal{A}_k \delta \mathcal{Z}_k + \mathcal{B}_k \delta u_k
$%\end{equation}
, where $\mathcal{A}_k$ and $\mathcal{B}_k$ are the linearization of the information-state dynamics about the nominal trajectory and $\delta \mathcal{Z}_k= (\delta z_k,\delta z_{k-1},\cdots,\delta z_{k-q},\delta u_{k-1},\cdots,\delta u_{k-q})$, where $\delta z_k = z_k - \bar{z}_k$ and $\delta u_k = u_k - \bar{u}_k$ are the deviation from the nominal observation at time $k$.\\
\noindent\textit{\textbf{Backward pass -}}
Given the LTV parameters about the nominal system, the ILQR algorithm computes a local optimal control as:
% by solving the following discrete-time Riccati equation:
% \begin{align}
%    \nonumber \delta u_k
%     %&=R^{-1}f_{u_k}'(-v_{k+1}-V_{k+1}\delta x_{k+1})-\bar{u}_k \nonumber \\
%     &=R^{-1}\mathcal{B}_k\T(-v_{k+1}-V_{k+1}(\mathcal{A}_k\delta \mathcal{Z}_k + \mathcal{B}_k\delta u_k))-\bar{u}_k, %\nonumber
%     %&=-(R+\mathcal{B}\T_k V_{k+1}\mathcal{B}_k)^{-1}(R\bar{u}_k+\mathcal{B}_k\T v_{k+1}+\mathcal{B}_k\T V_{k+1}\mathcal{A}_k\delta \mathcal{Z}_k),
% \end{align}
% which can be written in the linear feedback form 
$\delta u_k =-\kappa_k-K_k\delta \mathcal{Z}_k$, where $\kappa_k=(R+\mathcal{B}_k\T V_{k+1}\mathcal{B}_k)^{-1}(R\bar{u}_k+\mathcal{B}_k\T v_{k+1})$ and $K_k=(R+\mathcal{B}_k\T V_{k+1}\mathcal{B}_k)^{-1}\mathcal{B}\T_k V_{k+1}\mathcal{A}_k$, and
% \begin{subequations}
\begin{align}
    \nonumber v_k &= l_{k}^{\mathcal{Z}}+\mathcal{A}_k\T v_{k+1}-\mathcal{A}_k\T V_{k+1}\mathcal{B}_k(R+\mathcal{B}_k\T V_{k+1}\mathcal{B}_k)^{-1} \\
    &\hspace{4.2cm}\cdot(\mathcal{B}_k\T v_{k+1}+R\bar{u}_k), \label{eq:vt}\\
    % V_k &= l_{k,ZZ}+\mathcal{A}_k\T (V_{k+1}^{-1}+\mathcal{B}_k R^{-1}\mathcal{B}_k\T)^{-1}\mathcal{A}_k\nonumber \\
    V_k &=l_{k}^{\mathcal{ZZ}}+\mathcal{A}_k\T V_{k+1}\mathcal{A}_k-\mathcal{A}_k\T V_{k+1}\mathcal{B}_k(R+\mathcal{B}_k\T V_{k+1}\mathcal{B}_k)^{-1} \label{eq:Vt} \nonumber \\
    &\hspace{4.5cm}\cdot \mathcal{B}_k\T V_{k+1}\mathcal{A}_k,
\end{align}
% \end{subequations}
with the terminal conditions $v_N =\frac{\partial c_N}{\partial \mathcal{Z}}|_{\mathcal{Z}_N}$ and $V_N =\nabla ^2_{\mathcal{ZZ}}c_N|_{\mathcal{Z}_N}$. To compute $v_k$ and $V_k$, we need their corresponding values in the future, \textit{i.e.} $v_{k+1}$ and $V_{k+1}$. Given the terminal conditions from the forward pass and the local LTV model parameters $(\mathcal{A}_k, \mathcal{B}_k)$, we can do a backward-in-time sweep to compute the sequence $v_k$ and $V_k$, and thus, the corresponding optimal control for that trajectory. \\
\textit{\textbf{Update trajectory -}}
Given the gains from the backward pass, we can update the nominal control sequence as 
$%$\begin{align*}
    \bar{u}_k^{(i+1)} = \bar{u}_k^{(i)} + \kappa_k + K_k(\mathcal{Z}_k^{(i+1)}-\mathcal{Z}_k^{(i)}),%\\
    \mathcal{Z}^{(i+1)}_0 = \mathcal{Z}^{(i)}_0.
$%\end{align*}
~We reiterate the process till $R\bar{u}_k+\mathcal{B}_k\T v_{k+1}\approx 0$.

Albeit, in theory, the ILQR is easily generalized to the partially observed case, the key difficulty is to find the information-state LTV system parameters $\mathcal{A}_k, \mathcal{B}_k$. In general, we cannot evaluate these system parameters analytically, and thus, we need to evaluate them in a data-based fashion. 
%This section details the algorithm for open-loop trajectory design using POD-iLQR. The iLQR is a sequential quadratic programming-based approach to the solution of nonlinear optimal control problems that iteratively linearizes the dynamics and quadratizes the cost function around the current estimate of the optimal control sequence, solves an LQR problem to get an improved estimate of the control sequence, and proceeds recursively till one satisfies the necessary conditions. The advantage of iLQR is that the equations involved are given explicitly in terms of the linearized time varying (LTV) dynamics. 
Fortunately, this may be accomplished using our previously proposed Linear Time Varying-ARMA (LTV-ARMA) identification algorithm \cite{systemid_acc2023} which we recount briefly below.

%\vspace{-1mm}
\subsection{Information-State LTV System Identification using ARMA}
Let the nominal state be denoted by  $\bar{x}_k$ and the deviations from the nominal state as $\delta x_k$, which can be modeled as the following LTV system linearized around the nominal trajectory as: $ \delta x_k = A_{k-1} \delta x_{k-1} + B_{k-1} \delta u_{k-1}, ~ \delta z_k = C_k \delta x_k.$
We have the following result (see \cite{systemid_acc2023} for proof).

%\vspace{-1mm}
\begin{proposition}\label{p:ARMA}
An ARMA model of the order $q$ given by: $\delta z_{k} = \alpha_{k-1}  \delta z_{k-1}+\cdots+\alpha_{k-q}   \delta z_{k-q}   + \beta_{k-1} \delta u_{k-1}+\cdots+ \beta_{k-q}   \delta u_{k-q},$
exactly fits the LTV system given above if matrix $O^q = \begin{bmatrix} A_{k-q}\T...A_{k-2}\T C_{k-1}\T, & \cdots, & A_{k-q}\T C_{k-q+1}\T, & C_{k-q}\T \end{bmatrix}\T$ is full column rank. The exact parameters that match the LTV system can then be written as:
%\vspace{-1mm}
\begin{align}
    \nonumber [\alpha_{k-1} ~|~ & \alpha_{k-2} ~| \cdots|~\alpha_{k-q}] =  C_{k}A_{k-1}...A_{k-q}O^{q^+}, \\
    \nonumber [\beta_{k-1} ~|~ & \beta_{k-2} ~| \cdots|~\beta_{k-q}] = - C_{k}A_{k-1}...A_{k-q}O^{q^+} G^q \\
   \nonumber & + \begin{bmatrix}
    C_{k}B_{k-1} & C_{k}A_{k-1}B_{k-2} ~ \cdots ~ C_{k}A_{k-1}...B_{k-q}
    \end{bmatrix},
\end{align}

where matrix $G^q =$ 
\begin{align}
   \nonumber \begin{bmatrix}
  0 & C_{k-1}B_{k-2} & \cdots  & \cdots  & C_{k-1}A_{k-2}...B_{k-q}  \\ 
  0 & 0  & C_{k-2}B_{k-3}  & \cdots  &   C_{k-2}A_{k-3}...B_{k-q} \\ 
  \vdots & \vdots &  \ddots & \cdots  & \vdots  \\
  0&0 &0 & \cdots  & C_{k-q+1} B_{k-q} \\ 
  0&0 &0 & \cdots  & 0  \end{bmatrix}.
\end{align}
% $[\alpha_{k-1} ~|~ \alpha_{k-2} ~| \cdots|~\alpha_{k-q}] =  CA^{q}O^{q^+}$
% $[\beta_{k-1} ~|~ \beta_{k-2} ~| \cdots|~\beta_{k-q}] = \begin{bmatrix} CB & CAB & \cdots & CA^{q-1}B & CA^qB \end{bmatrix}- CA^{q}O^{q^+} G^q$.
\end{proposition}
% \begin{proof}
% Finally, substituting for $\delta Z_k^q$ and $\delta U_k^q$ gives the exact analytical solution for the ARMA parameters in terms of linear system matrices $A, B$ and $C$. 
% \end{proof}
% \vspace{1mm}
Notice that there always exists an exact fit for the ARMA model for any $q>\bar{q}$, if there exists a number $\bar{q}$ for which the matrix $O^{\bar{q}}$ is full column rank. Using this result, we can write linearized models in terms of ARMA parameters at each step along the nominal trajectory.

% This can also be understood as the capability to get the entire information-state from the past few observations and control inputs.
% \begin{remark}
% % The value of the number $q$ does not change at each time-step for the physical/mechanical systems examples we have tried.
% For the examples we have tried, we observed that the $q$ value does not change at each time-step and most likely would not change for physical/mechanical systems.
% \end{remark}

% \begin{corollary}\label{corollary2}
% For the case of mechanical systems with all the position/DOFs as the output feedback, the minimum value for $q$ would be $q=2$, which would allow for the exact fit for the ARMA model to be with only 2 past observations.
% \end{corollary}

%%%%%%%%%%%%%%%%%%%%%%%%%%%%%%%%%%%%%%%%%%%%%%%%%%%%%%%%%%%%%%%%%%%%%%%%%%%%%%%%%%%%%%%%%%%%%%%%%%%%%%%%%%%%%%%%%%%%%%%%%%%%%%%%%%%%%%%%%%%%%%%% 
%\vspace{1mm}
%\subsection{Linear Time-Varying System Identification using ARMA}
% \vspace{-1mm}
\label{sec_ls}
% Closed-loop control design in step 2 of D2C requires the knowledge of the linearized system parameters $A_{k}$ and $B_{k}$ for $0 \leq t \leq T-1$. 
% The standard least square method is used to estimate the linear parameters from input-output experiment data. 
Next, we show how to find the ARMA parameters from a suitable least squares problem. We perturb the linear system about the nominal trajectory and use the standard least square method to estimate the system parameters $\alpha_{k-i}$ and  $\beta_{k-i}$ for $i = 1, \cdots, q$ from:
% \begin{multline}
% \label{est_sys}
$ \delta z_{k}^{(j)} = \alpha_{k-1} \delta z_{k-1}^{(j)} +\cdots+\alpha_{k-q} \delta z_{k-q}^{(j)} + \beta_{k-1} \delta u_{k-1}^{(j)}+\cdots+ \beta_{k-q} \delta u_{k-q}^{(j)},
$
% \end{multline} 
where $\delta u_{k}^{(j)} \sim \mathcal{N}(0,\sigma I)$ is the control input perturbation at step $k$ for the $j$\textsuperscript{th} simulation.
% $\delta z_{k}^{(j)}$ is the observed output with the control input perturbation vector $\delta u_{k}^{(j)} \sim \mathcal{N}(0,\sigma I)$ that we feed to the system at step $t$ for the $j$\textsuperscript{th} simulation.
% , where $j = 1, \cdots, N$ for a total of $N$ simulations. All the perturbations are zero-mean, i.i.d, Gaussian noise with the covariance matrix. The covariance $\sigma$ is a $o(u)$ small value selected by the user. 
% $\delta x_{k+1}^{(n)}$ denotes the deviation of the output state vector from the nominal state after propagating for one step.
% After running $N$ simulations for each step and collecting the data, we can write the linear mapping between input-output perturbation as:
The final linear relation between input-output perturbation and system parameters can be written as
% \begin{align}
% \label{sysmat}
% \nonumber
$
\mathbb{Z} = [\alpha_{k-1} ~|~ \alpha_{k-2} ~| \cdots|~\alpha_{k-q} ~|~ \beta_{k-1} ~|~ \beta_{k-2} ~| \cdots|~\beta_{k-q}] \mathbb{X},
$
% \end{align}
and can be solved using the standard least square method as:
$[\alpha_{k-1} ~| \cdots|~\alpha_{k-q} ~|~ \beta_{k-1} ~| \cdots|~\beta_{k-q}]  = \mathbb{Z} \mathbb{X}^{\dagger},$ where $\mathbb{X}^{\dagger}$ denotes the pseudo-inverse of some matrix $\mathbb{X}$.
Please check \cite{Wang_ICRA_2021} for more details.

\begin{table*}[h!]
% \vspace{-2mm}
% \centering
%     $
\begin{align}\label{tab}
    \setcounter{MaxMatrixCols}{20}
    \underbrace{
    \begin{bmatrix} 
    \delta z_{k} \\ \delta z_{k-1} \\ \delta z_{k-2} \\ \vdots \\ \delta z_{k-q+1} \\ \hline \delta u_{k-1} \\ \delta u_{k-2} \\ \delta u_{k-3} \\ \vdots \\ \delta u_{k-q+1} 
    \end{bmatrix}}_{\delta \mathcal{Z}_k} = 
    \underbrace{\begin{bmatrix} 
    \alpha_{k-1} & \alpha_{k-2} & \cdots  &  \alpha_{k-q+1}  & \alpha_{k-q} & \vline &  \beta_{k-2} & \beta_{k-3} & \cdots  &  \beta_{k-q+1}  
    & \beta_{k-q}    \\  
    1 & 0 & \cdots & 0 & 0 & \vline & 0 & 0 & \cdots & 0 
    & 0    \\  
    0 & 1 & \cdots & 0 & 0 & \vline & 0 & 0 & \cdots & 0 
    & 0    \\  
    \vdots & & \ddots &  & \vdots & \vline & \vdots &  & \ddots & \vdots
    & 0    \\
    0 & 0 & \cdots & 1 & 0 & \vline & 0 & 0 & \cdots & 0 
    & 0    \\
    \hline
    0 & 0 & \cdots & 0 & 0 & \vline & 0 & 0 & \cdots & 0 
    & 0    \\
    0 & 0 & \cdots & 0 & 0 & \vline & 1 & 0 & \cdots & 0 
    & 0     \\  
     0 & 0 & \cdots & 0 & 0 & \vline & 0 & 1 & \cdots & 0 
    & 0     \\  
    \vdots & & \ddots &  & \vdots & \vline & \vdots &  & \ddots & & \vdots    \\
    0 & 0 & \cdots & 0 & 0 & \vline & 0 & 0 & \cdots & 1
    & 0
    \end{bmatrix} }_{\mathcal{A}_{k-1}}
    \underbrace{\begin{bmatrix} \delta z_{k-1} \\ \delta z_{k-2} \\ \vdots \\ \delta z_{k-q+1}  \\ \delta z_{k-q} \\ \hline  \delta u_{k-2} \\ \delta u_{k-3} \\   \vdots \\ \delta u_{k-q+1} \\ \delta u_{k-q} \end{bmatrix}}_{\delta \mathcal{Z}_{k-1}} 
    + \underbrace{\begin{bmatrix} \beta_{k-1} \\ 0 \\ \vdots \\ 0 \\ 0 \\ \hline 1 \\ 0  \\ \vdots \\  0 \\0 \end{bmatrix}}_{\mathcal{B}_{k-1}} \delta u_{k-1} 
    % + \underbrace{\begin{bmatrix} \gamma_{k-1} \\ 0 \\ \vdots \\ 0 \\ 0 \\ \hline 0 \\ 0  \\ \vdots \\  0 \\0 \end{bmatrix}}_{\mathcal{D}_{k-1}} w_{k-1}
\end{align}
    % $
%\vspace{-4mm}
\end{table*}
%%%%%%%%%%%%%%%%%%%%%%%%%%%%%%%%%%%%%%%%%%%%%%%%%%%%%%%%%%%%%%%%%%%%%%%%%%%%%%%%%%%%%%%%%%%%%%%%%%%%%%%%%%%%
%\subsection{LTV System Dynamics in Information State}
After identifying the system parameters $\alpha_{k-1},\cdots,\alpha_{k-q}$ and $\beta_{k-1},\cdots,\beta_{k-q}$ for $k = \{0, \cdots, N \}$, we write the perturbation LTV system in the information-state as given in Eq.~(\ref{tab}), which is written with the observation model as:
% where $\gamma_{k-1}$ is given as:
% $    \gamma_{k-1} = \beta_{k-1} + \beta_{k-2} + \cdots + \beta_{k-q},$ as the random noise sequence $\{w_{k-1}, w_{k-2}, \cdots, w_{k-q}\}$ is independent and assuming the noise to enter the same channel as the control input. This is based on the idea that if the output $z_k$ depends on the last $q$ control inputs, then it would also depend on the last $q$ disturbance terms. 
% Finally, we write the LTV system in Information State as:
$%$\begin{align}
%\vspace{-1mm}
    \delta \mathcal{Z}_k = \mathcal{A}_{k-1} \delta \mathcal{Z}_{k-1}+ \mathcal{B}_{k-1} \delta u_{k-1}, ~\delta {z}_k = \mathcal{C}_k \delta \mathcal{Z}_k, %\label{eq:LinDyn_Z}
$%\end{align}
with $\mathcal{C}_k = [I_{n_z} ~ 0]$, where $n_z$ is the number of measured outputs. Then we use the identified $\mathcal{A}_{k-1}$, $\mathcal{B}_{k-1}$ to find the optimal nominal trajectory using iLQR steps given earlier with alternating between backward and forward passes until convergence.
\section{Partially-Observed Decoupled Data-based Control (POD2C) Algorithm}\label{s:POD2c}
%\vspace{-1mm}
We now propose an extension to the so-called decoupled data-based control (D2C) algorithm \cite{wang2021decoupled,cdc_soc}. The D2C algorithm is a highly data-efficient Reinforcement Learning (RL) method that has shown to be much superior to state-of-the-art RL algorithms in terms of data efficiency and training stability while having better performance. %In the following, we detail the extension that allows for the generation of the closed-loop output feedback policy, i.e., the feedback as a function of the past few observations. The main observation for the extension is that the partially observed case might be treated similarly to the fully observed case by noting that the information-state comprises of past $q$ observations and control inputs as detailed earlier.
% \begin{remark}
% Please refer to Section II for ARMA model design which allows for writing information-state with finite past observations and control inputs.
% \end{remark}
% , $Z_k = (z_k,z_{k-1},\cdots,z_{k-q+1},u_{k-1},\cdots,u_{k-q+1})$.
%Notice that assuming sufficient smoothness, any feedback law for the information-state problem can be represented as $\pi_k(\mathcal{Z}_k) = \bar{u}_k + K_k \delta \mathcal{Z}_k+ S_k (\delta \mathcal{Z}_k)$, where $\bar{u}_k$ is the nominal control sequence arising from $\pi_k(\cdot)$, $K_k$ is the optimal linear feedback term (\cref{eq:MPK}) and $S_k(\cdot)$ are the higher order terms in the feedback law. 
The POD2C algorithm proposes a 2 step procedure to approximate the solution for the control problem in the presence of process as well as sensor noise. We first solve a noiseless open-loop optimization problem to find an optimal control sequence, $\bar{u}^*_k$ via the POD-iLQR scheme.
Then, an LQG controller is synthesized for the LTV perturbation model described in the information-state 
% $\delta \mathcal{Z}_k= {\mathcal{A}_{k-1}} \delta \mathcal{Z}_{k-1}+ {\mathcal{B}_{k-1}} \delta u_{k-1}$.
: $\delta \mathcal{Z}_k= {\mathcal{A}_{k-1}} \delta \mathcal{Z}_{k-1}+ {\mathcal{B}_{k-1}} \delta u_{k-1}+ {\mathcal{D}_{k-1}} w_{k-1}$ and measurement model: $\delta \mathcal{Y}_{k}  = \delta \mathcal{Z}_k+ v_k$, where $w_k$ and $v_k$ represent the process and measurement noise. The final control input to the system is $u_k= \bar{u}^*_k - K_k \delta \hat{\mathcal{Z}}_k$, where $K_k$ is the time varying feedback gain and $\hat{\mathcal{Z}}_k$ is the estimate of the information-state which is calculated as $\delta \hat{\mathcal{Z}}_k = \mathcal{A}_{k-1} \delta \hat{\mathcal{Z}}_{k-1} + {\mathcal{B}_{k-1}} \delta u_{k-1} + L_{k} (\delta \mathcal{Y}_{k} - \mathcal{A}_{k-1} \delta \hat{\mathcal{Z}}_{k-1} - {\mathcal{B}_{k-1}} \delta u_{k-1})$, where $L_k$ is the observer gain.
\section{Empirical Results}
\begin{figure}[!htbp]
\begin{multicols}{4}
    % \hspace{-3cm}    
    %   \subfloat[Cart-Pole]{\includegraphics[width=1\linewidth]{figures/init_cartpole.PNG}}    
    %   \subfloat[15-link Swimmer]{\includegraphics[width=1.005\linewidth]{figures/init_s15.PNG}}
    \subfloat{\includegraphics[width=1.2\linewidth]{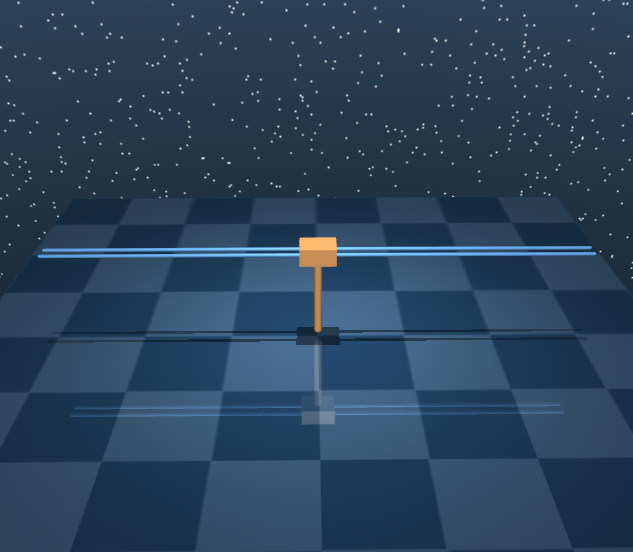}}
      \subfloat{\includegraphics[width=1.205\linewidth]{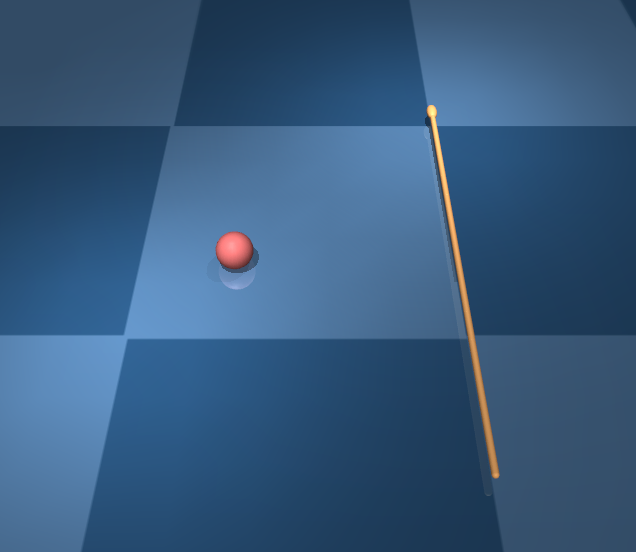}}
      \subfloat{\includegraphics[width=1.21\linewidth]{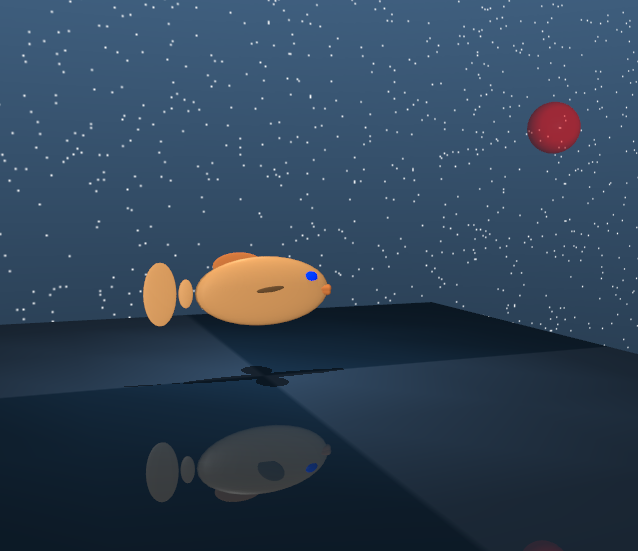}}
      \subfloat{\includegraphics[width=1.205\linewidth]{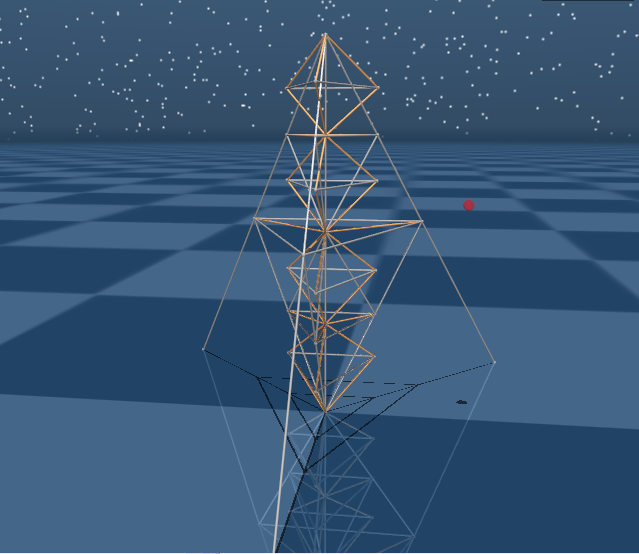}}
      \newline
      \subfloat{\includegraphics[width=1.2\linewidth]{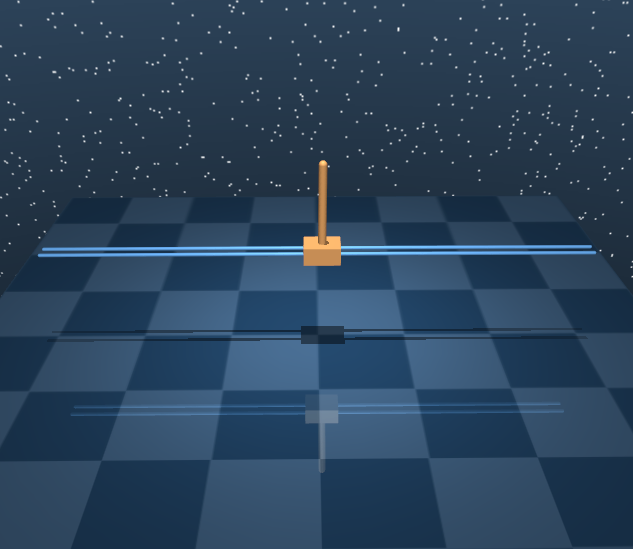}}    
       \subfloat{\includegraphics[width=1.204\linewidth]{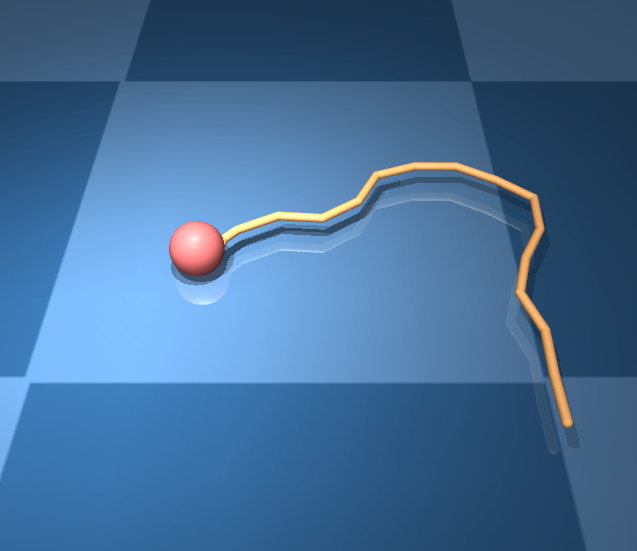}}
       \subfloat{\includegraphics[width=1.208\linewidth]{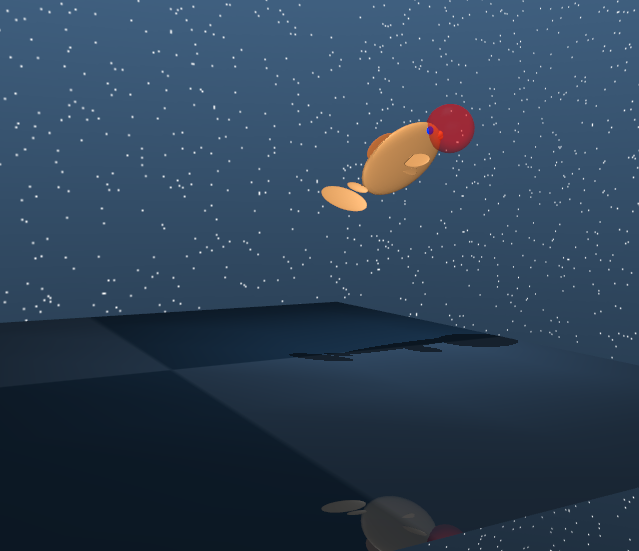}}
       \subfloat{\includegraphics[width=1.202\linewidth]{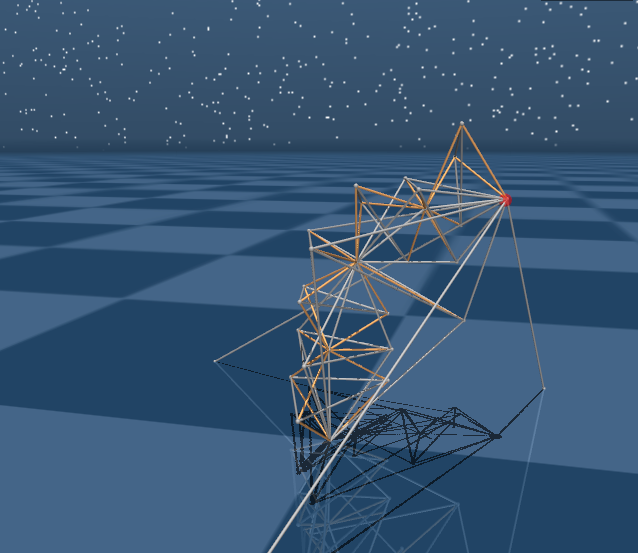}}
\end{multicols}
%\vspace{-3mm}
\caption{\small Models simulated in MuJoCo in their initial and final states.}
\label{figinit}
\end{figure}

\begin{figure}[!htbp]
\centering   
\subfloat[Initial state]{\includegraphics[width=0.4\linewidth]{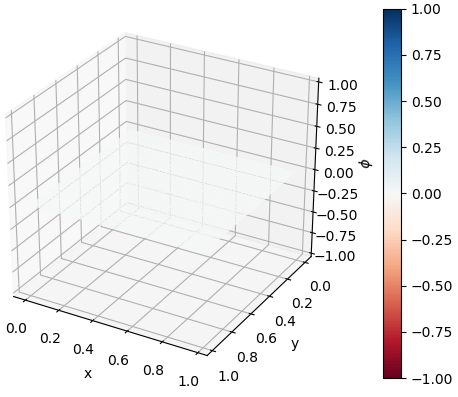}}
\subfloat[Final state]{\includegraphics[width=0.4\linewidth]{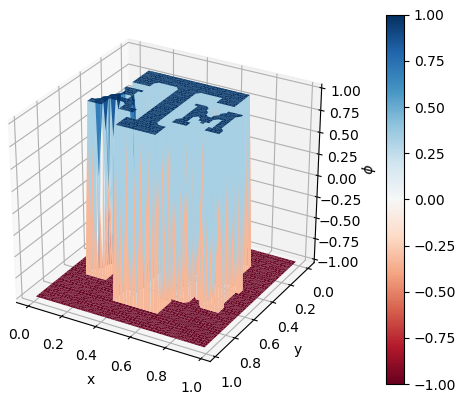}} 
\caption{\small High dimensional, complex, and partially observed nature of the problem governed by Allen-Cahn PDE (2500 states, 16 outputs)}
\label{figfinal1}
\end{figure}

%\vspace{-1mm}
We use a black box MuJoCo simulator \cite{todorov2012mujoco} to design the nominal trajectory and closed-loop feedback policy for the system with initial configurations given in Figure~\ref{figinit}.\\
\noindent \textbf{Cart-Pole:} The classic robotics model \cite{Wang_ICRA_2021}.\\
\noindent \textbf{15-link Swimmer:} The system has 17 DOF and together with their rates and controls are torques on the 14 joints.\\
% %\vspace{-1mm}
\noindent \textbf{Fish:} The fish has 13 DOF and 6 control channels with torques on the fin and tail joints.\\
    \noindent \textbf{T2D1 Robotic Arm:} 
The tensegrity model consists of 33 bars and 46 strings with controls as tensions in the strings.\\
\noindent \textbf{1D Viscous Burgers PDE:}
$\frac{\partial u}{\partial t}+u \frac{\partial u}{\partial x}=\nu \frac{\partial^2 u}{\partial x^2}$. We take external control inputs on the boundaries corresponding to blowing/suction such that $u(0,t)=U_1(t)$; $u(L,t)=U_2(t)$. Given an initial velocity profile $u(x,0) = \sin (\pi (x-1))$, the control task is to reach the goal state $u(x,T) = -0.50$ for $x = [0:0.02:2]$ with two observers added at both ends of the boundary. \\
\noindent \textbf{Allen-Cahn PDE:}
$\frac{\partial \phi}{\partial t}= -M(\frac{\partial F}{\partial\phi}-\gamma\nabla^{2}\phi)$. We adopt the following general form of energy density function \textit{F}: $F(\phi;T,h) = \phi^{4}+T\phi^{2}+h\phi$, where $\phi(x,t)\in [-1,1]$ is the order parameter, and $T(x,t)$ and $h(x,t)$ are external control inputs.
A periodic boundary condition is applied, and the control scheme is such that all grid points converging to the same goal state $\phi_k =1$ or $-1$ are fed the same control inputs.
Given a uniform initial condition $\phi_0=0$, the control task is to reach the custom goal-state as shown in Fig.~\ref{figfinal1}. 
\begin{figure*}[!htbp]
\begin{multicols}{4}
    \hspace{.3cm}    
      \subfloat[Cart-Pole]{\includegraphics[width=1.07\linewidth]{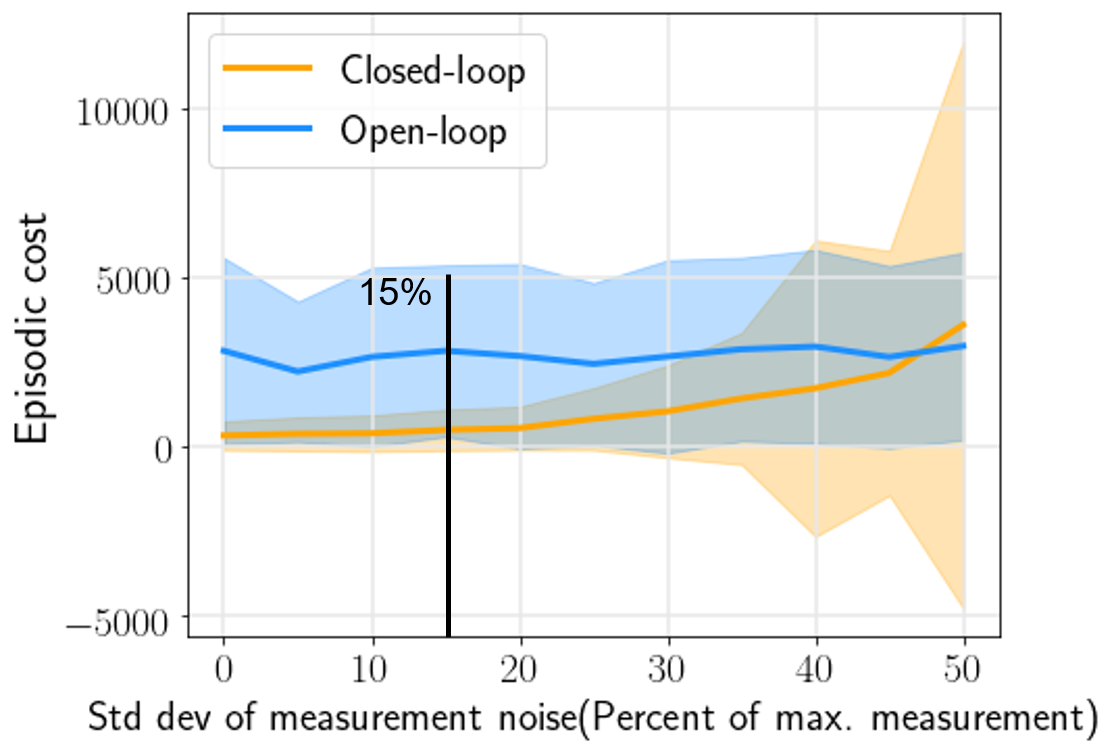}}    
      \subfloat[15-link Swimmer]{\includegraphics[width=1.05\linewidth]{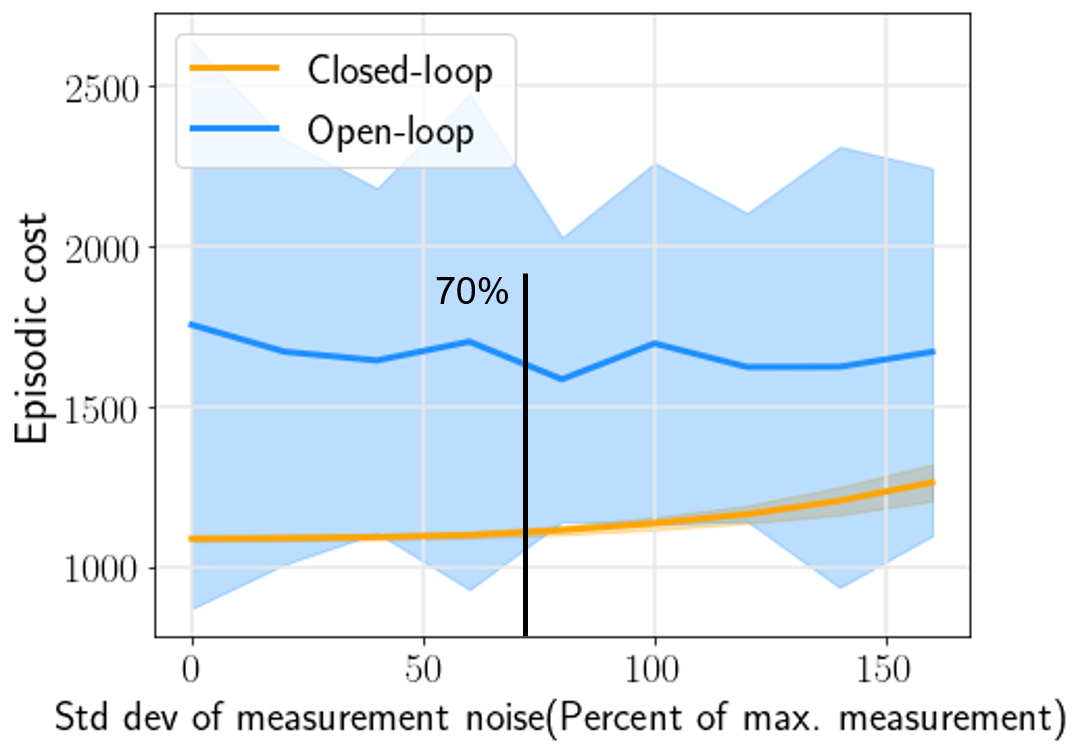}}
      \subfloat[Fish]{\includegraphics[width=1.05 \linewidth]{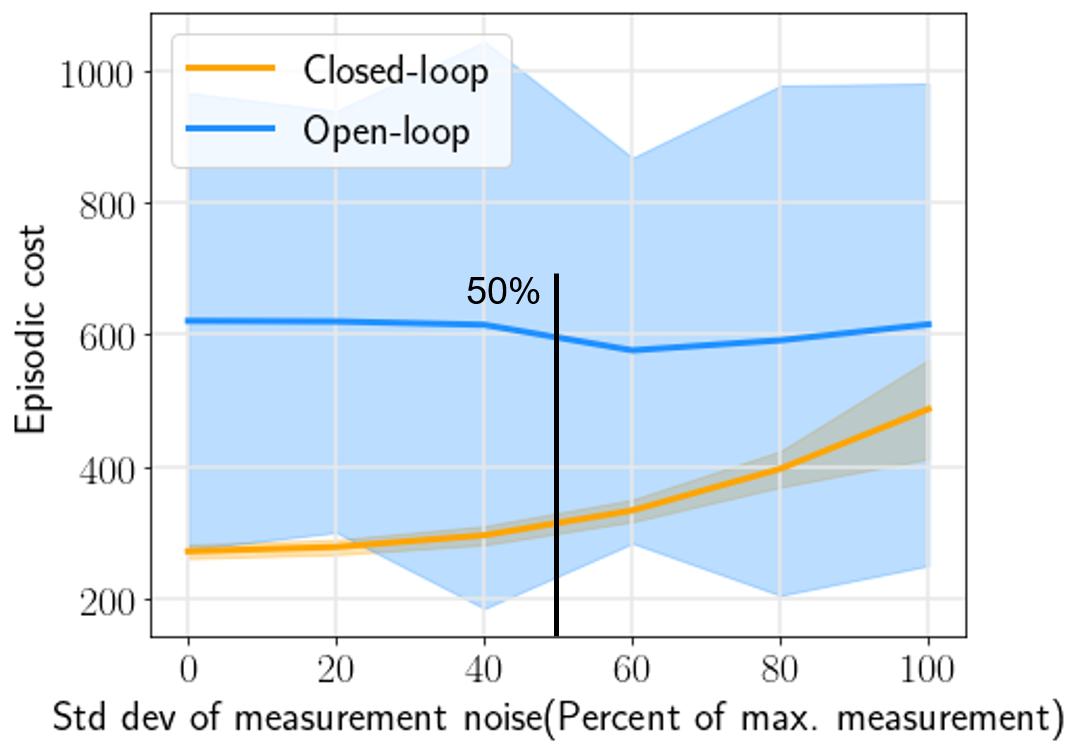}}
      \subfloat[T2D1 Robotic Arm]{\includegraphics[width=1.1\linewidth]{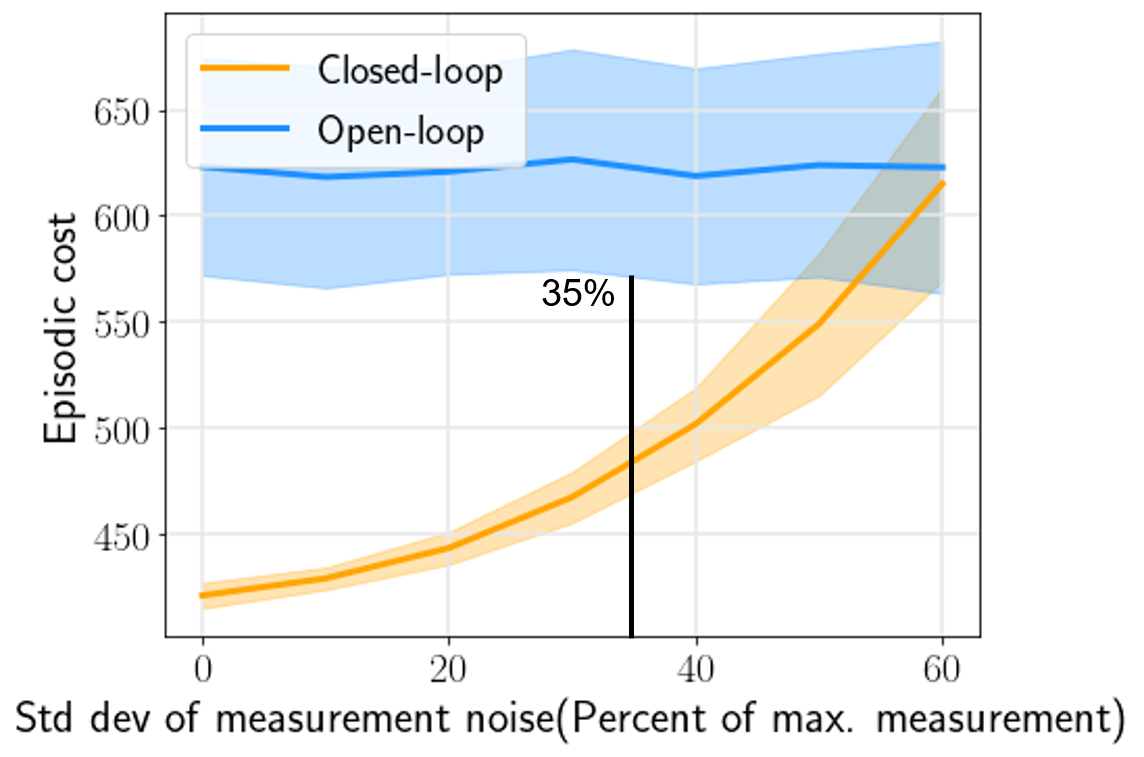}}
\end{multicols}
%\vspace{-3mm}
\caption{Averaged episodic reward vs measurement noise level for fixed 10\% process noise with LQG as closed-loop feedback}
%\vspace{-3mm}
\label{figmeasnoise}
\end{figure*}
\begin{figure*}[!htbp]
\begin{multicols}{4}
    \hspace{.4cm}    
      \subfloat[Cart-Pole]{\includegraphics[width=1.04\linewidth]{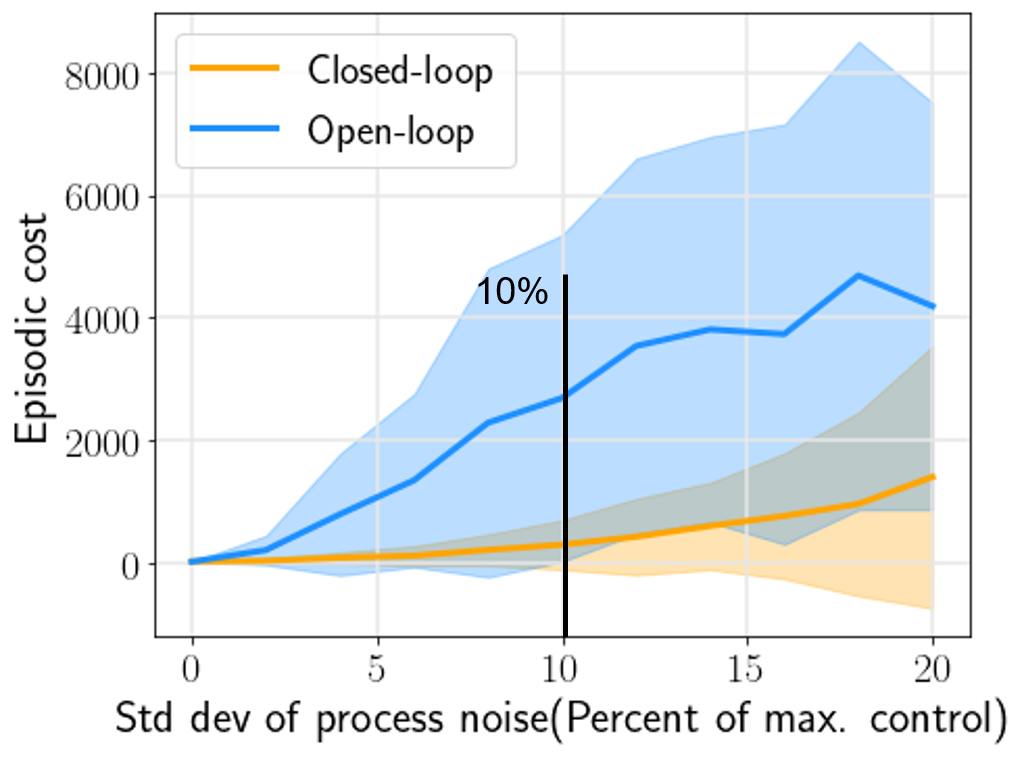}}    
      \subfloat[15-link Swimmer]{\includegraphics[width=1.04\linewidth]{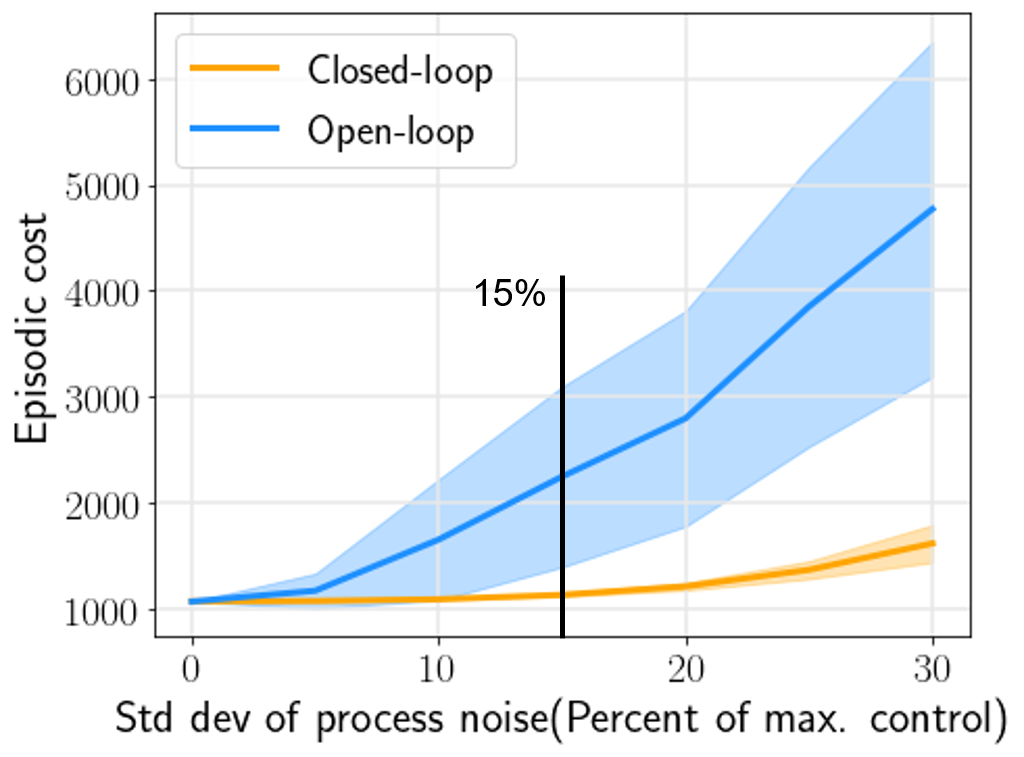}}
      \subfloat[Fish]{\includegraphics[width=1\linewidth]{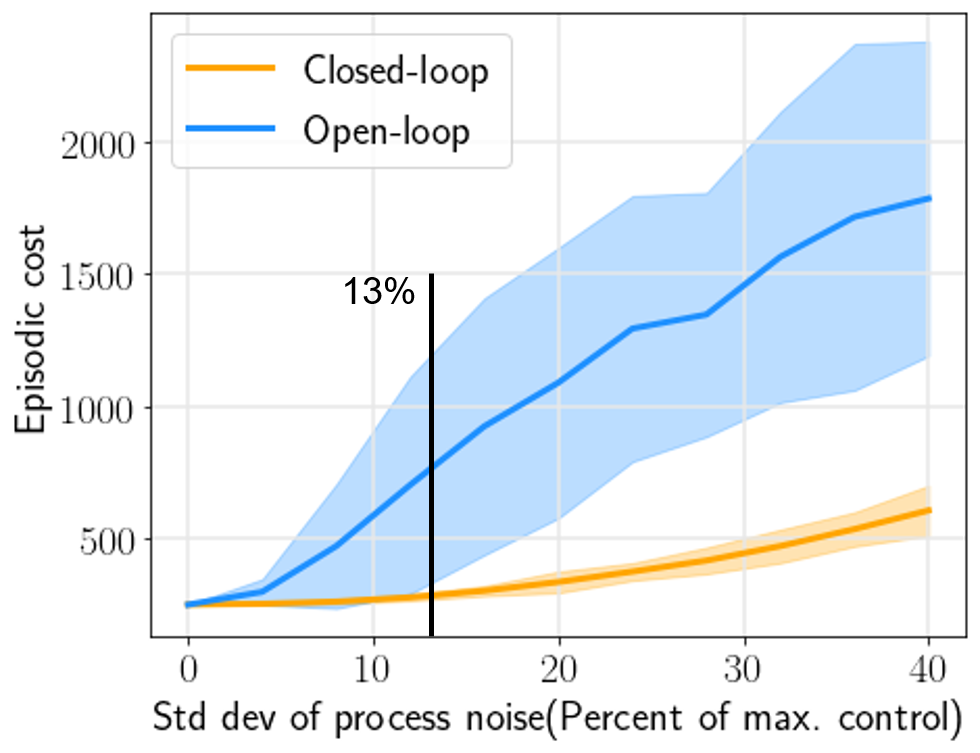}}
      \subfloat[T2D1 Robotic Arm]{\includegraphics[width=1.04\linewidth]{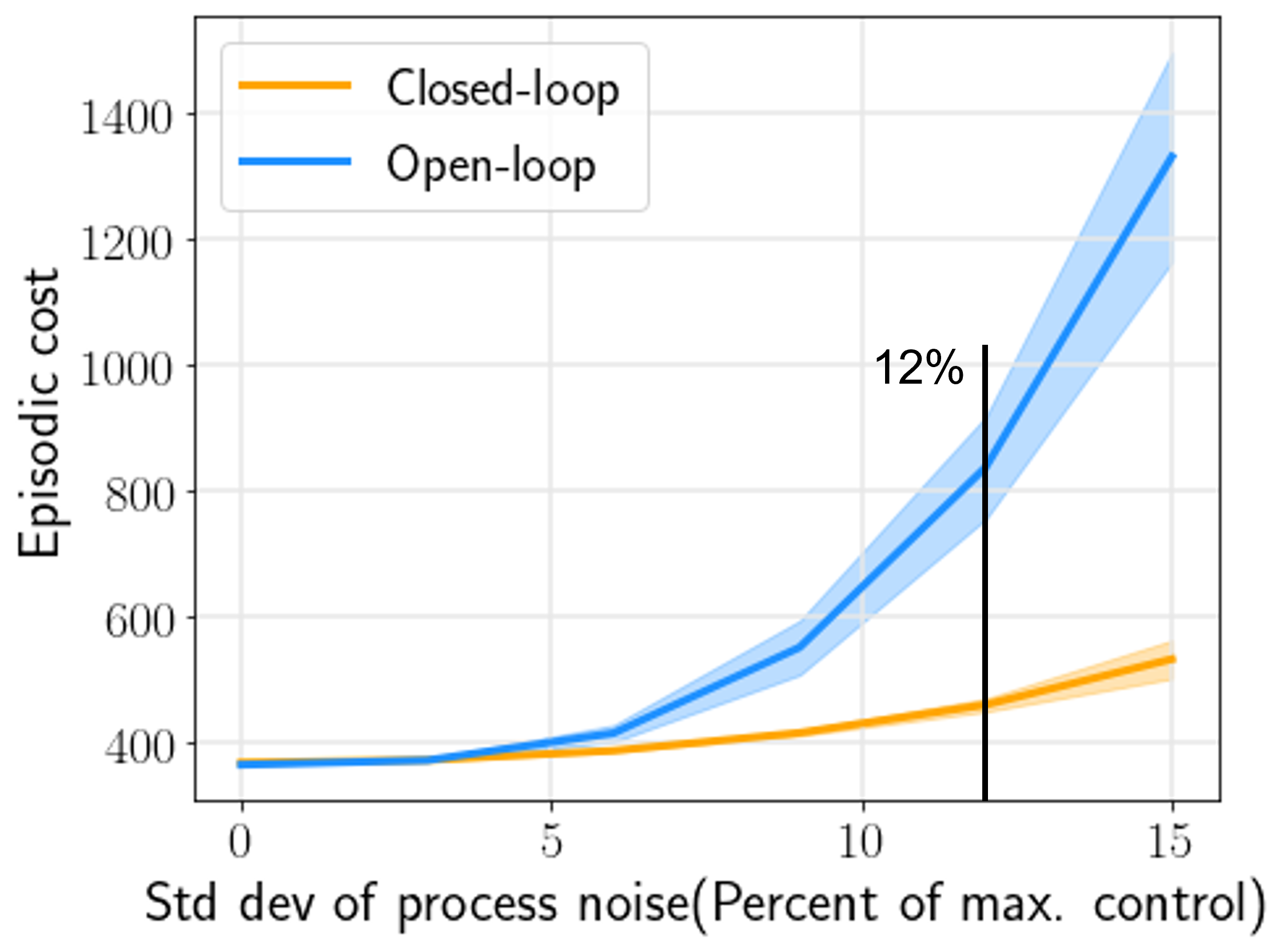}}
\end{multicols}
%\vspace{-3mm}
\caption{Averaged episodic reward vs process noise level for fixed 10\% measurement noise with LQG as closed-loop feedback}
%\caption{Averaged episodic reward vs measurement noise level for a fixed value of process noise with LQG as closed-loop feedback}
%\vspace{-5mm}
\label{figprocessnoise}
\end{figure*}
% The rule of thumb for measurement selection is that we only measure the positions that contain the most information and avoid redundant information.

Table~\ref{arma para} provides the time comparison to generate optimal open-loop trajectory between the gradient descent \cite{Wang_ICRA_2021} and proposed POD-iLQR. The output number in Table~\ref{arma para} represents the minimum number of measurements needed to fit a good $q\textsuperscript{th}$ order ARMA model. 
We observed that as the dimension of the system increases, the relative number of measurements required decreases. 
The cart’s position and pole-angle are used as feedback for the cart-pole model.
% The 15-link swimmer and fish are both high-dimensional multi-body robots in a fluid environment.
The 15-link swimmer requires only angular positions of every other joint and the fish needs only the angles of the fin and tail joints. The $T_2D_1$ arm is a soft-robotic arm where the node-positions of 8 (out of 25) node points were used.\\
% \vspace{-3mm}

% \vspace{-3mm}
\begin{table}[!htbp]
\caption{\small Comparison results between Gradient Descent \cite{Wang_ICRA_2021} and POD-iLQR (*Gradient descent failed to converge)} \label{arma para}
% \vspace{-2mm}
\begin{center}
% \vspace{0.1in}
\setlength{\tabcolsep}{1mm}{
\begin{tabular}{|c|c|c|c|c|c|c|}
\hline\textbf{}&\textbf{Grad. Des.}& \textbf{POD-iLQR} &&  && \\
\textbf{System}&\textbf{Total Time}& \textbf{Total Time}&\textbf{Iteration}& $q$ &\textbf{States}&\textbf{Outputs} \\
&\textbf{(sec.)}&\textbf{(sec.)}&\textbf{Number}& & \textbf{} & \textbf{}\\
\hline
Cart-pole& 15 & 0.8&40 & 2 & 4& 2\\
\hline
Swimmer & 376736.72 & 2110.0&100 & 5&34 & 10\\
% Swimmer&  &  & & & \\
\hline
Fish & * & 1720.5&200 &2&27 & 11\\
\hline
Robotic Arm &* & 355.8&40 &3&150 & 24\\
% Robotic Arm & &  && & \\
\hline
Burgers Eqn. &2763.4 & 38.2&10 &2&100 & 2\\
\hline
Allen-Cahn Eqn. &21078.18 & 47.78&10 &2&2500 & 16\\
\hline
\end{tabular}
}
\end{center}
% \vspace{-2mm}
\end{table}

%\vspace{1mm}
% The results are also given for viscous Burger's equation with two observers added at both ends of the boundary.
\textbf{Nominal trajectory design with POD-iLQR:}
% As described in Sec. \ref{s:POD2c}, we obtain the nominal trajectory from POD-iLQR training. 
% As shown in Fig. \ref{figcostconvergence}, the cart-pole training converged quickly in one second consisting of 40 iterations. The 15-link swimmer and the fish take $\approx 2000$ seconds with 100 iterations and $\approx 2300$ seconds with 200 iterations respectively. 
The 15-link swimmer and the fish take more iterations to converge as they have higher non-linearity due to the fluid-structure interaction in the swimming motion. However, the training is far more time-efficient compared with the first-order gradient descent method used in our prior work \cite{Wang_ICRA_2021}. The T2D1 arm system also takes less time and iterations
% $\approx 500$ seconds with 25 iterations 
to converge despite the high dimensionality of the model and limited outputs. The iteration numbers given in Table~\ref{arma para} show smooth and efficient convergence, even for systems with high non-linearity and high dimensionality. The results are obtained using MatLab and MuJoCo on a Ryzen 3700 PC. The most time-consuming procedure is running simulations to collect data for fitting the ARMA model, which is in serial for now but can be in parallel as the rollouts are independent of each other and further improve the time efficiency and have the potential for real-time operation.
% The obtained nominal control trajectory is then used as the open-loop nominal control policy in the following steps.

\textbf{Robustness to measurement and process noise:}
Figure~\ref{figmeasnoise} shows the plots for the episodic cost with the variation in the measurement noise. The figure compares the open-loop and the closed-loop control policies under different measurement noise levels, while the process noise standard deviation is set to 10\% of the maximal nominal control. The measurement noise level on the x-axis is the percentage of the measurement noise standard deviation w.r.t. the maximal nominal measurement. Note that both the measurement and process noise is added as zero-mean Gaussian i.i.d. noise to all measurement and control channels at each step. 

%As the measurement noise does not influence the open-loop, the open-loop cost curves are almost flat with invariant variance. 
%The closed-loop cost has a significantly smaller mean and variance than the open-loop cost, which proves the robustness to measurement noise of the closed-loop policy. 
Note that the closed-loop policy can successfully finish the task with smaller noise levels than what is indicated by the black threshold lines in the figure. 
% Also, the variance of the open-loop policy, as well as the variance of the closed-loop policy at zero measurement noise, come from the fixed 10\% process noise. %The spikes in the open-loop curves are due to numerical errors from the Monte-Carlo simulations.
% Notice that the mean episodic cost and the cost variance of the closed-loop policy, even with partial observation, is much smaller than that of the open-loop policy for all the examples for the entire noise level range.  
%\textbf{Robustness to process noise:}
% Figure~\ref{figprocessnoise} shows the plots for the episodic cost when we vary the process noise level along the x-axis with fixed 10\% measurement noise. 
Under the open-loop policy, the process noise drives the model off the nominal trajectory and results in high episodic cost, while the closed-loop feedback can help the system stay close to the nominal trajectory. This can be seen from the figure as the episodic cost mean and variance of the closed-loop policy is much smaller than the open-loop policy on the entire tested noise range, although both policies fail the task when the process noise becomes larger than what the black threshold line indicates. The threshold values of process and measurement noise are empirically calculated at which the system will go too far away from the nominal trajectory and the final goal can not be achieved. The above analysis regarding the performance of control policy under noise proves that the LQG closed-loop feedback wrapped around the nominal trajectory makes the full closed-loop policy robust to measurement and process noise.

\textbf{Comparison with a Direct RL Method:}
%Another angle to interpret the POD2C is that we use a linear ARMA model to fit the linearized dynamics which is then used in the iLQR open-loop training and LQG feedback design. By decoupling the deterministic open-loop optimization and the closed-loop feedback in stochastic systems, POD2C is efficient to obtain the control policy, thus feasible for complex models. 
In a direct RL method such as DDPG \cite{lillicrap2015continuous}, deep neural networks are used to represent the closed-loop control policy. In general, Direct RL methods require full state observation and it is not clear how to generalize them to partially observed problems. Nonetheless, we run the DDPG method on the fish example with the same information-state as found by POD2C. After training for approximately 20 hours, the DDPG algorithm failed to converge and the fish could not swim to the target.

%%%%%%%%%%%%%%%%%%%%%%%%%%%%%%%%%%%%%%%%%%%%%%%%%%%%%%%%%%%%%%%%%%%%%%%%%%%%%%%%%%%%%%%%%%%%%%%%%%%%%%%%%%%%
% \vspace{-1mm}
\section{Conclusions}
% \vspace{-1mm}
The paper proposed an optimal information-state approach to transform partially observed problems into fully observed optimal nonlinear control problems. We showed that the information-state based formulation is equivalent to the original system formulation with the partial nonlinear observation model. The paper further provides the conditions to meet the minimum principle in the information-state formulation. We then provide the algorithm that generated $q$\textsuperscript{th}-order ARMA models using the input-output data to model LTV systems in the information-state which further allowed designing the optimal nominal trajectory using iLQR with partial state observations. 
% The constructed LTV system in the information-state is shown to exactly match the Markov parameters of the underlying linear state-space model.
Finally, the paper presented a decoupled data-based approach to control complex robotic systems by designing the closed-loop feedback law with partial state observations.
%A unique kind of LQG algorithm is then formulated where all the states are already available from the information-state representation. 
Empirical results are shown for complex robotic systems, including challenging cases of hard-to-model soft contact constraints, dynamic fluid-structure interactions, and partial differential equations, under motion as well as sensing uncertainty. 
In our opinion, the POD2C approach is highly efficient for RL in partially observed problems, however, questions regarding optimality under noise shall be explored in future work. 
% Future work will explore partially observed non-smooth motion planning scenarios and another direction would be problems that require information-seeking behavior which considers the dual effect of control. We conjecture that a hybrid of our POD2C approach and dual effect Gaussian belief space planning might be useful in this regard.

% \begin{align}
%   {\bar{u}_{k}^{n}} &= {\bar{u}_k^{n-1}} + \alpha \kappa_k + K_{k} ({ \bar{z}^{n-1}_k} - {\bar{z}_k^{n-1}}),\\
% {\bar{u}_{k}^{*}} &= {\bar{u}_{k}^{n}},~after~convergence. 
% \end{align}

% \clearpage
% \vspace{-2mm}
\bibliographystyle{IEEEtran}
\bibliography{Bib,Refs-D2C-Proposal}

\end{document}